%% file: main.tex
\documentclass{article} 
\usepackage{iclr2027_conference,times}

\input{math_commands.tex}

\usepackage{hyperref}
\usepackage{url}
\usepackage{graphicx}
\usepackage{wrapfig}
\usepackage{pifont}
\usepackage{booktabs}
\usepackage{multicol}
\usepackage{multirow}
\usepackage[table]{xcolor}
\usepackage{xspace}
\usepackage{algorithm}
\usepackage{algpseudocode}
\usepackage{amsmath,amssymb}

\definecolor{bestcell}{RGB}{226,239,218}
\definecolor{secondcell}{RGB}{252,228,236}

\newcommand{\icon}{\raisebox{-2pt}{\includegraphics[width=1.em]{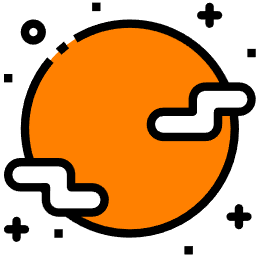}}\xspace}

\title{\icon HALO: Enhancing Time Series Generation via Hyperspherical Latents and Masked AutoregRessive Modeling}

\iclrfinalcopy
\author{
\textbf{Chunyi Hou,
Xiangfei Qiu,
Hanyin Cheng},
\textbf{Yutong Li,
Bin Yang} \\
School of Data Science and Engineering,
East China Normal University, Shanghai, China}
\begin{document}

\maketitle
\lhead{Preprint}
\begin{abstract}
\input{Abstract/abstract_2}
\end{abstract}


\section{Introduction}
\input{Introduction/introduction}

\section{Related Works}
\input{Related_Works/related_works}

\section{Preliminaries}
\input{Preliminaries/preliminaries}

\section{Method}\label{sec:4}
\input{Method/method}

\section{Experiments}\label{sec:5}

\input{Experiments/experiments}

\section{Conclusions}
\input{Conclusions/conclusions}

\bibliography{iclr2027_conference}
\bibliographystyle{iclr2027_conference}

\clearpage

\appendix

\input{Appendix/appendix}

\end{document}

%% file: math_commands.tex
\usepackage{amsmath,amsfonts,bm}

\def\eqref#1{equation~\ref{#1}}

\def\1{\bm{1}}

\DeclareMathAlphabet{\mathsfit}{\encodingdefault}{\sfdefault}{m}{sl}
\SetMathAlphabet{\mathsfit}{bold}{\encodingdefault}{\sfdefault}{bx}{n}



%% file: Abstract/abstract_2.tex
Most existing time series generators rely on a two-stage modeling paradigm: \textbf{the first stage} learns discrete latent representations of time series; \textbf{the second stage} performs autoregressive modeling on these discrete latents through next token prediction.
However, this paradigm suffers from two stage-specific limitations: the first stage can lead to information loss when discretizing continuous time series, while the second stage is prone to error accumulation during autoregressive generation.
To address these limitations, our core idea is to perform generative modeling in a continuous latent space with a more efficient autoregressive framework.
We propose \underline{\textbf{HALO}}, which enhances time series generation via \underline{\textbf{H}}yperspheric\underline{\textbf{A}}l \underline{\textbf{L}}atents and Masked Aut\underline{\textbf{O}}regressive modeling to achieve this goal by tackling two key bottlenecks:
(1) variance and scale heterogeneity of continuous latent representations; 
(2) the difficulty of balancing generation efficiency with temporal correlation modeling.
HALO first introduces a hyperspherical VAE that constrains continuous latents to a fixed-radius hyperspherical shell, effectively stabilizing the numerical fluctuations of continuous latent representations.
Secondly, we develop a masked autoregressive model that balances parallel decoding and temporal correlation learning, substantially reducing the number of inference steps required for generation and improving generation stability.
Our extensive experiments demonstrate that HALO achieves state-of-the-art generation performance while offering significantly improved inference efficiency over existing advanced baselines.


%% file: Introduction/introduction.tex
Time series generation plays an important role in a wide range of real-world domains, including finance~\citep{financesurvey,financesurvey2,financesurvey3}, healthcare~\citep{healthcaresurvey,healthcaresurvey2}, and industrial fault diagnosis~\citep{diagnosisurvey,diagnosisurvey2}.
The success of deep generative models in natural language processing~\citep{languagesurvey} and computer vision~\citep{visionsurvey} has motivated their application to time series generation.
By synthesizing realistic time series, these approaches can help alleviate data scarcity in settings where access to real-world data is constrained by privacy concerns or the rarity of extreme events~\citep{zhang2026survey}.

Existing time generation methods~\citep{sdformer,timemar,msdformer,polarformer} generally adopt a \emph{two-stage modeling paradigm}.
\textbf{In the first stage}, a discrete tokenizer, such as VQ-VAE~\citep{vqvae}, is trained to tokenize time series into discrete latent representations.
\textbf{In the second stage}, an autoregressive model~\citep{gpt4} is trained, following the "\emph{next token prediction}" to model the conditional distributions of these discrete tokens~\citep{ntp}.
Although they have achieved promising results, they still suffer from several limitations.
Firstly, discrete tokenization is inherently misaligned with the continuous nature of time series, potentially causing the loss of fine-grained temporal information.
Secondly, autoregressive models rely on sequential token-by-token generation, which is prone to error accumulation and often leads to unstable and inefficient generation~\citep{exposurebias}.
To address these key limitations, we aim to perform generative modeling in a continuous latent space with a more efficient autoregressive framework, thereby mitigating information loss and improving generation stability and efficiency.
Nonetheless, achieving this goal needs to overcome the two key challenges:


\begin{figure}[t]
    \begin{center}
    \includegraphics[width=0.47\linewidth]{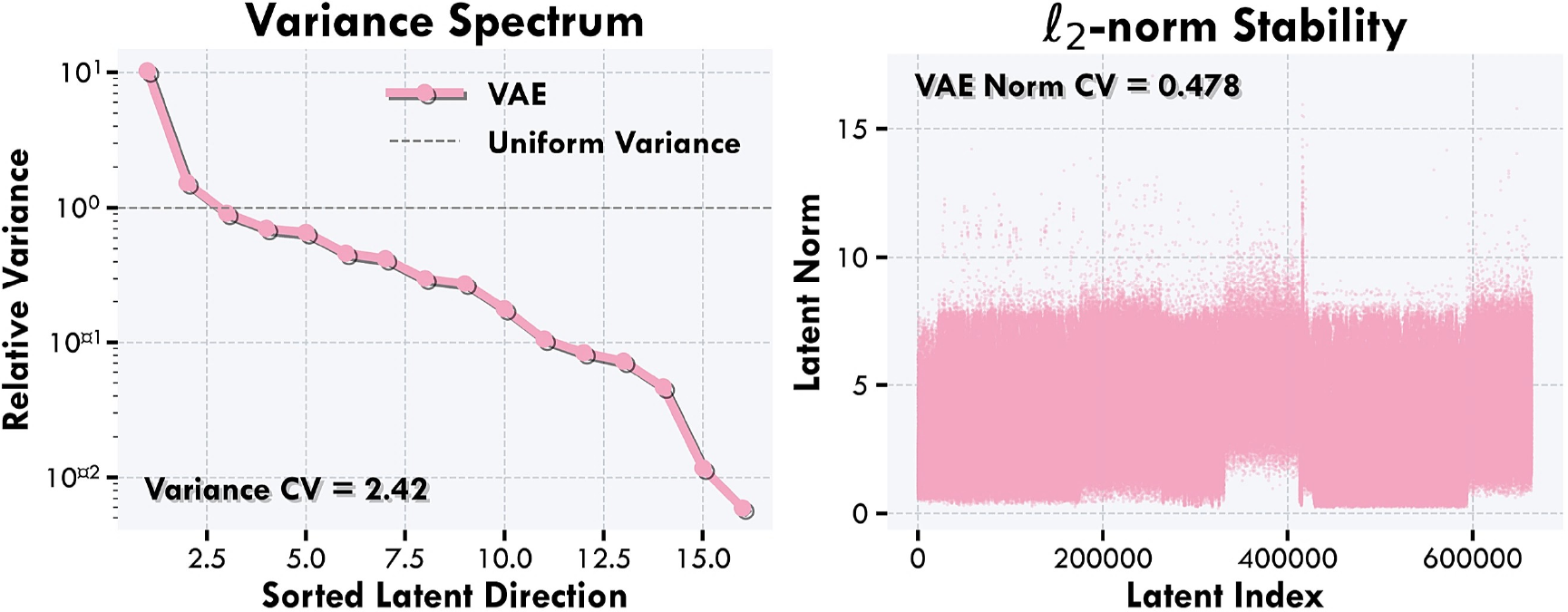}
    \hspace{0.04\linewidth}
    \includegraphics[width=0.47\linewidth]{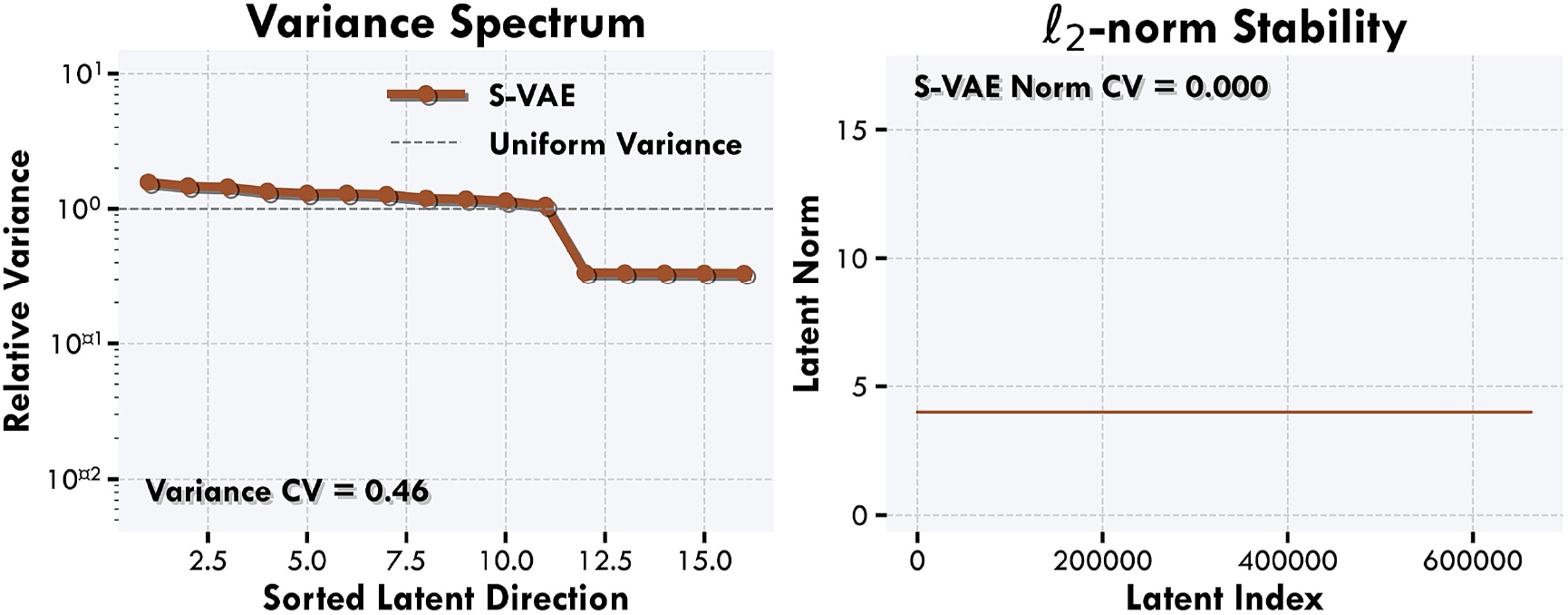}
    \end{center}
    {\hspace{0.21\linewidth}(a) VAE\hspace{0.43\linewidth}(b) S-VAE}
    \hspace{-18pt}
    \caption{\small Quantitative comparison of VAE~\citep{vae} and S-VAE (Ours) latents: relative variance spectra and token norms.
    Relative variance denotes covariance eigenvalues normalized by
    their mean, with dashed lines indicating uniform variance.
    CV denotes the coefficient of variation.
    S-VAE exhibits more balanced directional variances and nearly
    constant token norms.\label{fig:1}}
    \vspace{-18pt}
\end{figure}
\textbf{Challenge 1: Continuous tokens exhibit substantial variance and scale heterogeneity.}
When modeling non-stationary and temporally heterogeneous time series, continuous tokens without explicit constraints on their numerical range may exhibit substantial variance and scale heterogeneity.
As shown in Figure~\ref{fig:1}(a), we quantitatively analyze the continuous tokens produced by a Gaussian VAE~\citep{vae}.
Several clear observations can be drawn:
\textcolor{red}{\ding{172}} the normalized covariance eigenspectrum decays rapidly, indicating
that most of the latent variance is concentrated along only a few
principal directions;
\textcolor{red}{\ding{173}} the latent norms vary substantially across tokens, revealing considerable token-wise scale heterogeneity.
These valuable insights indicate that continuous tokens exhibit both directional variance imbalance and latent-wise scale variation.
Overall, these forms of continuous latents heterogeneity severely increase the learning and modeling difficulty of the second-stage autoregressive model.

\begin{wrapfigure}{r}{0.48\textwidth}
\vspace{-14pt}
    \begin{center}
    \includegraphics[width=1.\linewidth]{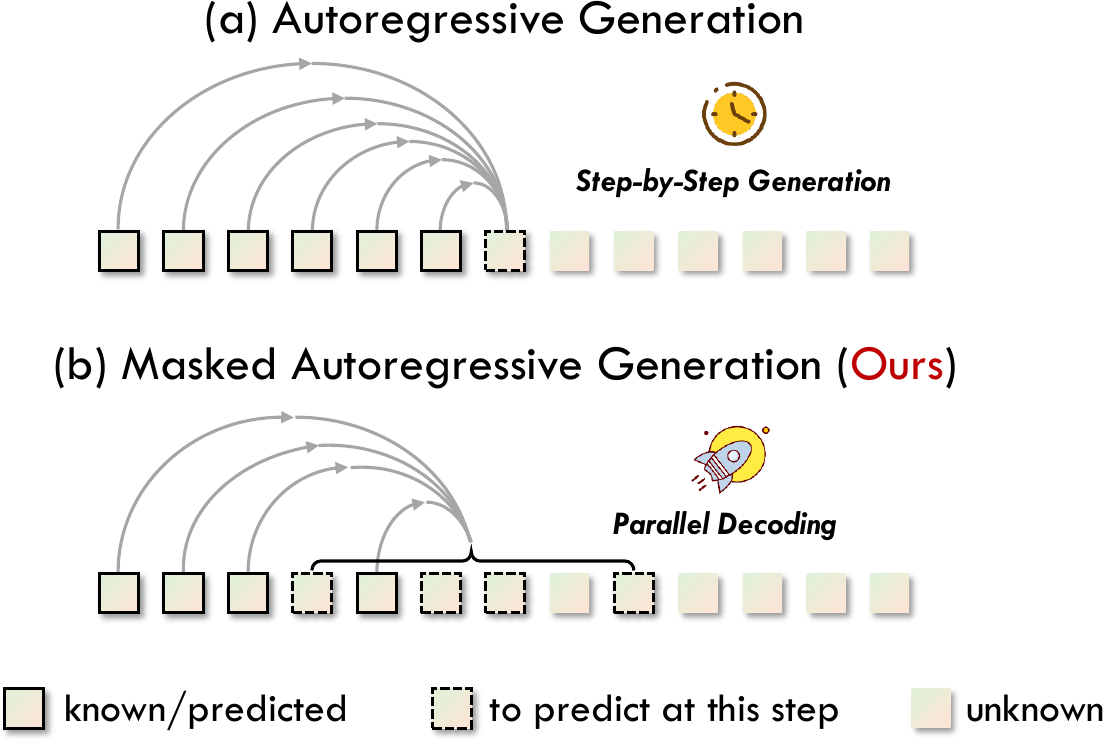} \\
    \end{center}
    \vspace{8pt}
    \caption{\small(a) \emph{Autoregressive generation} produces tokens sequentially, and is prone to unstable and inefficient.
    (b) \emph{Masked autoregressive generation} predicts unknown tokens conditioned on known tokens, achieving stable and efficient generation.\label{fig:2}}
    \vspace{-16pt}
\end{wrapfigure}
\textbf{Challenge 2: Efficient generation while preserving temporal correlation modeling.}
Standard autoregressive models explicitly model temporal dependencies by predicting each token conditioned on the previous ground-truth tokens~\citep{ntp}, as shown in Figure~\ref{fig:2}(a).
Such an architecture provides the model with a strong capability to model temporal correlations, while this sequential, token-by-token decoding process limits generation efficiency.
A straightforward and widely adopted alternative is to use MAE-style masked representation learning~\citep{mae,Visionts}, which enables more efficient generation through parallel token prediction.
Although this approach alleviates the inefficiency of token-by-token generation, it weakens the advantage of explicit temporal correlation modeling, resulting in suboptimal generation quality and fidelity~\citep{sdformer}.

Motivated by the aforementioned observations and reflections, we address the above challenges by introducing {\textbf{HALO}}, a novel generative framework, which enhances time series generation via \underline{\textbf{H}}yperspheric\underline{\textbf{A}}l \underline{\textbf{L}}atents and Masked Aut\underline{\textbf{O}}regressive modeling.
Unlike existing two-stage deep generative models for time series generation in a discrete representation space, HALO introduces a hyperspherical VAE (S-VAE) and masked autoregressive modeling to alleviate variance and scale heterogeneity of the continuous latent representations and balance efficiency and stability during autoregressive generation, respectively.
More specifically, to address \textbf{the first challenge}, we introduce an S-VAE based on the von Mises–Fisher distribution.
The S-VAE constrains continuous latent representations to a hypersphere with a fixed radius, making the continuous latents learned by the autoregressive model scale-invariant.
As illustrated in Figure~\ref{fig:1}(b), the latents produced by our S-VAE exhibit more stable scale dynamics and greater directional homogeneity, with their norms strictly constrained to the fixed radius of the hypersphere.
Therefore, the heterogeneities of continuous latents are substantially alleviated, reducing the difficulty of learning their distributions for the autoregressive model.
Furthermore, as an additional benefit, this design mitigates scale drift associated with variance and scale heterogeneity during autoregressive generation~\citep{LatentLM}.

On the other hand, to address \textbf{the second challenge}, inspired by recent advances in autoregressive models in visual domain~\citep{hmar,maskgit,mage}, we introduce masked autoregressive modeling.
Following this idea, we develop a more general autoregressive model that \emph{predicts unknown tokens conditioned on known tokens} to strike a balance between masked representation learning and unidirectional autoregressive models, as shown in Figure~\ref{fig:2}(b).
As a result, it is empowered with strong capabilities for both parallel decoding and temporal correlation learning.
These compact designs not only improve the model’s generative capability, but also significantly reduce the number of inference steps required for generating a single sample, thereby improving generation stability and efficiency.
Our contributions can be summarized as follows:
\begin{itemize}

\item 
We reflect the limitations of existing two-stage deep time series generative models, particularly the loss of fine-grained temporal information and error accumulation during autoregressive generation.
Therefore, we propose HALO, a novel deep generative framework that performs generative modeling in a continuous latent space with an efficient autoregressive framework, enabling the generation of high-quality and high-fidelity synthetic time series.

\item 
We design a novel S-VAE that constrains latents to lie on a fixed-radius hypersphere, effectively mitigating variance and scale heterogeneity in continuous latent representations. 
At the same time, this fundamentally alleviates the degradation in generation quality and fidelity caused by scale drift during autoregressive generation.
    
\item 
We develop a masked autoregressive model that predicts unknown tokens conditioned on known tokens to strikes a balance between parallel decoding and temporal correlation learning.
While preserving temporal awareness, it significantly reduces the number of inference steps required for generation, thereby improving generation efficiency and stability.

\item 
Extensive experiments demonstrate that HALO achieves state-of-the-art generation performance.
Moreover, compared with other strong baselines, HALO also demonstrates superior inference efficiency.
Our code and scripts are available at \href{}{here}.

\end{itemize}




%% file: Related_Works/related_works.tex
\subsection{Deep Generative Models for Time Series}
Recent years have witnessed rapid advances in deep generative modeling
for time series. 
Early studies such as TimeGAN~\citep{TIMEGAN} combine
supervised and adversarial objectives in a jointly learned embedding
space to preserve temporal dynamics. 
As an alternative to adversarial generation, TimeVAE~\citep{timevae} employs an interpretable encoder-decoder architecture that explicitly models characteristic
temporal components such as trends and seasonalities.
As generative modeling techniques continue to advance, two-stage deep generative models have continued to draw considerable attention from both researchers and practitioners~\citep{latentdiffusion,hdt,li2026sdflow}.
As a representative work, SDformer~\citep{sdformer} introduces the discrete-latent generative modeling paradigm to the time series generation community.
In the first stage, it trains a VQ-VAE~\citep{vqvae} to compress time series into a discrete latent space, while the second stage employs a GPT-style autoregressive model to learn the joint conditional probability distribution over the resulting discrete latent representations.
Recently, to better model multi-scale temporal dynamics in time series, TimeMAR~\citep{timemar} and MSDformer~\citep{msdformer} further advance this paradigm by extending VQ-VAE to multi-scale architectures.
Despite their effectiveness, these methods represent intrinsically continuous signals using a finite codebook, which introduces quantization error and limits the fidelity of fine-grained temporal variations.

\subsection{Autoregressive Modeling for Time Series Generation}

Beyond natural language processing, computer vision, and audio generation, autoregressive models have also been widely adopted in the time series generation.
They can be broadly categorized into two lines: GPT-style unidirectional autoregressive models and BERT-style bidirectional autoregressive models~\citep{mage}.
For unidirectional autoregressive models, representative methods such as SDformer~\citep{sdformer} and MSDformer~\citep{msdformer} adopt the next token prediction paradigm.
For instance, TimeMAR~\citep{timemar} builds a scale-wise autoregressive model upon multi-scale latent representations, enabling next-scale prediction.
PolarFormer~\citep{polarformer} introduces a polar coordinate decomposition technique to alleviate codebook collapse and training difficulties.
However, despite their strong temporal correlation learning capability, these unidirectional autoregressive models still face dilemmas in generation stability and efficiency.
In another line of research, bidirectional autoregressive models, such as SDformer-m~\citep{sdformer} and TimeVQVAE~\citep{timevqvae}, are driven by the cloze-style prediction paradigm.
Compared with unidirectional autoregressive models, they offer the advantage of parallel decoding, thereby substantially improving generation efficiency~\citep{mae}.
Yet, due to their weakened capability for temporal correlation modeling, they often suffer from suboptimal performance.
Moving beyond these two autoregressive paradigms, HALO develops a more general masked autoregressive model.
This model returns to the fundamental principle of autoregressive modeling by predicting unknown tokens conditioned on known tokens, while striking a balance between parallel decoding and temporal correlation learning to enable both efficient and high-quality time series generation.

%% file: Preliminaries/preliminaries.tex
\subsection{Definition}
\paragraph{Definition 3.1 (Multivariate Time Series)}
For the given multivariate time series $X\in\mathbb{R}^{N\times T}$, where $N$ represents the number of variates, and $T$ is the number of time steps.
HALO adopts a channel-independent strategy~\citep{patchtst} for robust modeling. 
Accordingly, for simplicity, we omit the variate dimension in the following formulation.

\paragraph{Definition 3.2 (VAE)}

The VAE~\citep{vae} is widely employed to compress input data into a lower-dimensional latent vector.
It consists of an encoder $q_\phi(z \mid x)$, which parameterizes the approximate posterior distribution over \(z\).
Symmetrically, it also includes a decoder $q_\psi(x \mid z)$, which reconstructs \(x\) from \(z\).
The standard VAE training objective is formulated by maximizing the evidence lower bound (ELBO):
\begin{equation}
    \mathcal{L}_{ELBO}= \mathbb{E}_{q_\phi(\mathbf{z}\mid \mathbf{x})}\left[\log p_\psi(\mathbf{x}\mid \mathbf{z})\right]-D_{\mathrm{KL}}\left(q_\phi(\mathbf{z}\mid \mathbf{x})\,\|\, p(\mathbf{z})\right).
\end{equation}
Both the prior \(p(z)\) and the approximate posterior \(q_\phi(z\mid x)\) are parameterized as Gaussian distributions with diagonal covariance matrices by default; the prior is the isotropic standard Normal $\mathcal{N}(\mathbf{0},\mathbf{I})$.
During training, the reparameterization trick is typically employed, i.e., \(z = \mu_\phi(x) + \sigma_\phi(x) \odot \epsilon\), where \(\epsilon \sim \mathcal{N}(\mathbf{0},\mathbf{I})\). 
By decoupling the stochastic sampling from the encoder parameters, this operation enables gradients to be backpropagated through \(z\) from the decoder to the encoder.
Under this diagonal Gaussian posterior, the encoder outputs data-dependent, dimension-wise scale \(\sigma_\phi(x)\), leading to heterogeneous variances across both latent dimensions and latents.
Moreover, such heterogeneous variances can be progressively amplified during autoregressive decoding, causing detrimental scale drift~\citep{LatentLM}.

\subsection{Problem Statement}
Given a training set \(\mathcal{D}=\{X^{(i)}\}_{i=1}^{D}\) consisting of \(D\) time series samples, the goal is to learn a parameterized distribution \(p_{\theta}(X)\) that approximates the underlying data distribution \(p_{\mathrm{data}}(X)\).
Once trained, the model can generate synthetic time series \(\hat{X} \sim p_{\theta}(\cdot)\).
The generated time series should faithfully preserve the temporal dynamics and statistical properties of the real data.

%% file: Method/method.tex
\subsection{Structure Overview}
\begin{figure}
    \centering
    \includegraphics[width=0.97\linewidth]{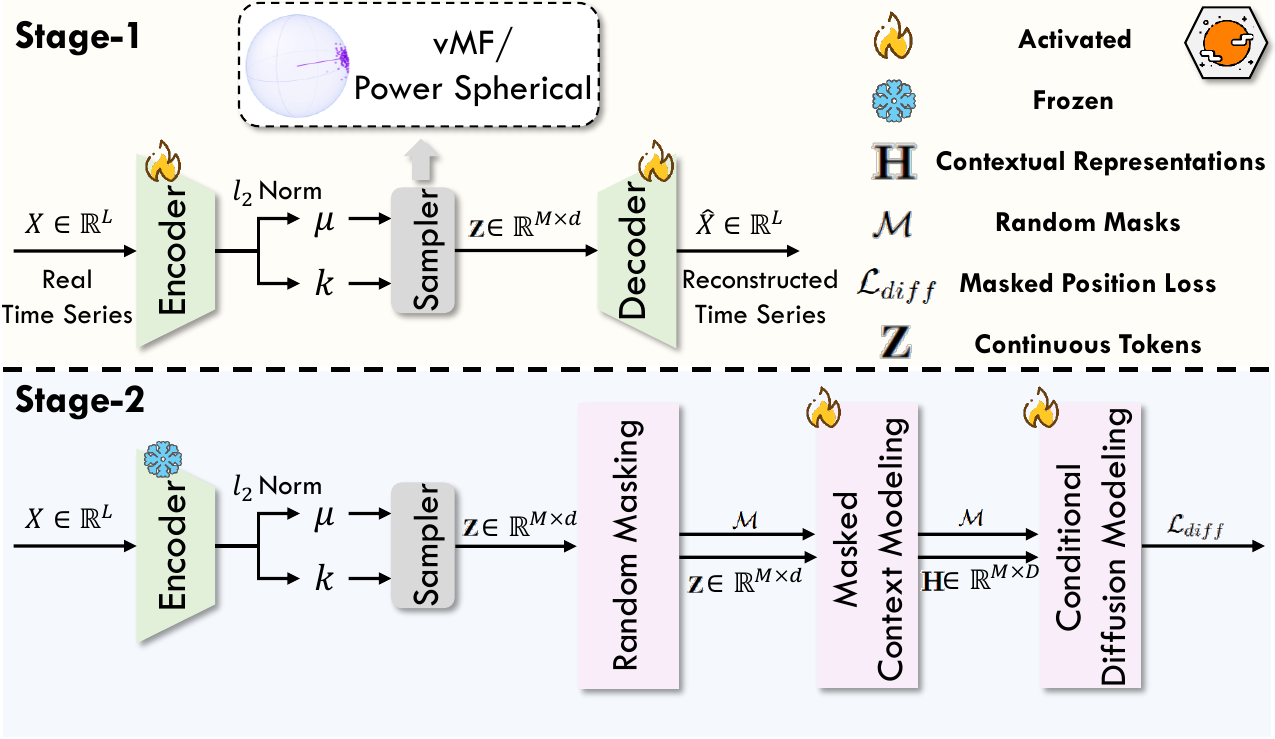}
    \caption{\small \icon Overview of the HALO framework.}
    \label{fig:3}
\end{figure}

HALO adopts a classical two-stage generative modeling paradigm as illustrated in Figure~\ref{fig:3}.
In the \textbf{first stage}, we train an S-VAE with a symmetric encoder–decoder architecture under a reconstruction objective.
First, the encoder of S-VAE compresses the real time series into continuous latent representations and models them with a von Mises–Fisher distribution, constraining the latents to a fixed-radius hypersphere and thereby alleviating variance and scale heterogeneity in continuous latents.
Next, given the latent representations produced by the encoder, the decoder reconstructs the time series from them.


In the \textbf{second stage}, we train a masked autoregressive model to learn the distribution of the continuous latent representations.
This model mainly includes three components: a random masking strategy, masked context modeling, and conditional diffusion modeling.
The masking strategy selects prediction targets, while masked context modeling derives conditioning representations from the visible tokens.
The conditional diffusion model then learns the distributions of the masked tokens conditioned on these representations.
This design balances generation efficiency and temporal correlation modeling while helping mitigate error accumulation.


\subsection{Stage 1: Hyperspherical VAE (S-VAE)}
\paragraph{Encoder}
For the encoder, following conventional practice, we progressively compress the real time series $X\in\mathbb{R}^{L}$ into low-dimensional latent representations ${h}\in\mathbb{R}^{M\times d}$, where $M=L/s$ is the number of latents, and $s$ represents the compression ratio.
Formally, it can be expressed as follows:
\begin{equation}
    {h}=\text{Encoder}(X).
\end{equation}
For the encoder architecture, we design a hybrid structure to more effectively capture temporal characteristics.
Unlike existing methods~\citep{sdformer,timemar}, we combine convolutional neural networks with Transformer blocks, enabling the encoder to preserve strong local feature extraction while leveraging the global modeling capability of Transformers.
This complementary design enables the proposed encoder to learn more expressive latents.


\paragraph{von Mises--Fisher Distribution}
After obtaining a sequence of continuous latents ${h}$ produced by the encoder, 
we constrain them to the unit hypersphere to mitigate heterogeneous variance 
in the continuous latent representations.
Specifically, we first employ a linear projection layer to output a direction, and a concentration for each latent. 
The direction is subsequently $\ell_2$-normalized 
to obtain the mean direction $\boldsymbol{\mu}\in\mathbb{S}^{d-1}$, while the 
concentration $\kappa\in\mathbb{R}_{\geq 0}$ is constrained to be non-negative.
Then, based on $\boldsymbol{\mu}$ and $\kappa$, we parameterize the approximate posterior 
using a von Mises--Fisher distribution~\citep{vfm}:
\begin{equation}
q_{\phi}(\mathbf{u}\mid\mathbf{x})
=
C_d(\kappa)
\exp\left(
\kappa\boldsymbol{\mu}^{\top}\mathbf{u}
\right),
\quad
\mathbf{u}\in\mathbb{S}^{d-1},
\end{equation}
where
$
C_d(\kappa)
=
\frac{\kappa^{\frac{d}{2}-1}}
{(2\pi)^{\frac{d}{2}}
I_{\frac{d}{2}-1}(\kappa)}
$
is the normalization constant, and $I_{\nu}(\cdot)$ denotes the modified Bessel 
function of the first kind of order $\nu$.
The \(\boldsymbol{\mu}\) controls the preferred direction, while the concentration \(\kappa\) determines the degree of concentration around that direction.

After parameterizing the directional posterior with the von Mises--Fisher distribution, we sample a direction \(\mathbf{u}\sim q_\phi(\mathbf{u}\mid\mathbf{x})\).
We adopt a uniform prior \(p(\mathbf{u})=\mathrm{Unif}(\mathbb{S}^{d-1})\) over the hypersphere, and transform the sampled direction to the fixed-radius latent representations through \(\mathbf{z}=R\mathbf{u}\).
By fixing the radial component, all latent representations share the same norm, thus eliminating variance and scale heterogeneity in the continuous latents.
Moreover, this geometric constraint reduces the degrees of freedom of the latent distribution, simplifying the distribution learning in the second-stage training.

\paragraph{Decoder}
Given the hyperspherical latent representations $\mathbf{Z}=[\textbf{z}_1,\textbf{z}_2,\ldots,\textbf{z}_M]$, the decoder maps it back to the original time-series space to reconstruct the input sequence:
\begin{equation}
    \hat{X}=\text{Decoder}(\mathbf{Z}),\quad\textbf{z}=R\textbf{u}.
\end{equation}
To this end, the decoder defines the conditional likelihood \(p_\psi(X\mid \mathbf{Z})\), which constitutes the ELBO of the S-VAE:
\begin{equation}\label{eq:5}
\begin{aligned}
\mathcal{L}_{{S-VAE}}
=
\mathbb{E}_{q_{\phi}(\mathbf{U}\mid{X})}
[\log p_{\psi}(X\mid \mathbf{Z}=R\mathbf{U})]-D_{\mathrm{KL}}
(q_{\phi}(\mathbf{U}\mid X)\|p(\mathbf{U})),\quad \mathbf{U}\in(S^{d-1})^M,
\end{aligned}
\end{equation}
where the first term related to the reconstruction, encouraging the decoder to reconstruct the input time series from the sampled latents, and the second term regularizes the approximate posterior toward the prior distribution on the hypersphere that promoting a well-structured latent space.

\paragraph{Learning Objective}
The overall training objective includes the $\mathcal{L}_{ELBO}$. 
In addition, we introduce a spectral loss \(\mathcal{L}_{\mathrm{FFT}}\), which computes an \(\ell_1\) loss in the frequency domain to encourage the S-VAE to faithfully preserve periodic and seasonal components:
\begin{equation}
    \mathcal{L}_{Total}=-\mathcal{L}_{S-VAE}+\alpha\mathcal{L}_{FFT},\quad
    \mathcal{L}_{FFT}=\|\mathcal{F}(\hat{X})-\mathcal{F}({X})\|_{1},
\end{equation}
where $\alpha$ is the hyperparameter, and $\mathcal{F}(\cdot)$ represents the Fast Fourier Transform.



\subsection{Stage 2: Mask Autoregressive Modeling}
\paragraph{Random Masking Strategy}

To learn latent distributions, we
construct masked prediction tasks over the latent sequences produced
by the frozen S-VAE under varying visible contexts, following the random masking paradigm~\citep{mage,maskgit}. 
Let
$\mathbf{Z}=[\mathbf{z}_1,\textbf{z}_2,\ldots,\textbf{z}_M]\in\mathbb{R}^{M\times d}$
denote a latent sequence.
For each mini-batch, we sample a masking ratio $r$ from a truncated
Gaussian distribution :
\begin{equation}
    r \sim \operatorname{TruncNormal}_{[r_{\min},1]}
    (\mu=1,\sigma^2=0.25^2),
    \qquad
    m = \lceil rM \rceil,
\end{equation}
where $r_{\min}$ controls the minimum masking ratio and $m$ is the
number of masked tokens.

Furthermore, for each sequence, we independently sample a uniformly random
permutation $\pi$ of the position indices $\{1,\ldots,M\}$.
The masked and visible position sets are defined as
\begin{equation}
    \mathcal{M}=\{\pi_1,\ldots,\pi_m\},
    \qquad
    \mathcal{V}=\{1,\ldots,M\}\setminus\mathcal{M}.
\end{equation}
This procedure uniformly selects $m$ distinct positions for masking,
while preserving the original temporal positions of the latent tokens.
The visible tokens $\mathbf{Z}_{\mathcal{V}}$ provide contextual
information, while the masked tokens $\mathbf{Z}_{\mathcal{M}}$ serve
as prediction targets. 
By varying both the masking ratio and the
masked positions, the model learns to predict missing latent tokens
from different amounts and configurations of visible context.



\paragraph{Masked Context Modeling}
To capture temporal correlations across tokens,
we employ a Transformer encoder-decoder to construct contextual
representations for the masked positions.
For the visible tokens $\mathbf{Z}_{\mathcal{V}}$,
the encoder uses bidirectional self-attention to model
relationships among different temporal positions, including
dependencies between distant tokens.
Learnable temporal positional embeddings retain the ordering information
of the latent sequence, while the domain label $\textbf{C}$ provides
domain-specific prompts.
Classifier-free guidance (CFG) can also be adopted to enable unconditional generation.

The decoder combines the encoded visible context with learnable
mask tokens at the missing positions.
Each masked position can attend to visible tokens from both
earlier and later temporal locations, allowing its contextual
representation to incorporate information from the surrounding
sequence as well as distant temporal context:
\begin{equation}
    \mathbf{H}
    =
    f_{\theta}
    \left(
    \mathbf{Z}_{\mathcal{V}},\mathcal{M},\textbf{C}
    \right),
\end{equation}
where $f_{\theta}$ denotes the context modeling network, and
$\mathbf{H}\in\mathbb{R}^{M\times D}$ contains the contextual
representations.
For each $i\in\mathcal{M}$, $\mathbf{H}_i$ summarizes the
temporal context relevant to the missing latent token and
serves as the condition for subsequent diffusion modeling.
This enables the conditional prediction of each masked token
to account for temporal correlations within the latent sequence.

\paragraph{Conditional Diffusion Modeling}

To model the distribution of each masked latent token $p_{\omega}(\mathbf{Z}_i\mid\mathbf{H}_i)$, we employ a lightweight conditional diffusion head, given the contextual representations $\mathbf{H}$, for $i\in\mathcal{M}$.
The contextual representation $\mathbf{H}_i$ carries temporal
dependencies extracted from the visible sequence, allowing
the diffusion head to predict the missing latent token
conditioned on its temporal context.

During training, we perturb each target latent token
$\mathbf{Z}_i$ with Gaussian noise:
\begin{equation}
    \mathbf{Z}_i^{(t)}
    =
    \sqrt{\bar{\alpha}_t}\,\mathbf{Z}_i
    +
    \sqrt{1-\bar{\alpha}_t}\,\boldsymbol{\epsilon}_i,
    \qquad
    \boldsymbol{\epsilon}_i\sim\mathcal{N}(\mathbf{0},\mathbf{I}),
\end{equation}
where $t$ denotes the diffusion timestep and
$\bar{\alpha}_t$ is determined by the noise schedule.
A lightweight MLP
$\boldsymbol{\epsilon}_{\omega}
(\mathbf{Z}_i^{(t)},t,\mathbf{H}_i)$
predicts the injected noise, while the timestep and contextual
information are integrated into noise prediction network through AdaLN~\citep{dit}.

We optimize the noise prediction objective over the masked
positions:
\begin{equation}
    \mathcal{L}_{{diff}}
    =
    \mathbb{E}\left[
    \frac{1}{|\mathcal{M}|d}
    \sum_{i\in\mathcal{M}}
    \left\|
    \boldsymbol{\epsilon}_i
    -
    \boldsymbol{\epsilon}_{\omega}
    \left(
    \mathbf{Z}_i^{(t)},t,\mathbf{H}_i
    \right)
    \right\|_2^2
    \right],
\end{equation}
where the expectation is taken over the training data,
random masks, diffusion timesteps, and Gaussian noise.
The diffusion head and the context modeling network are
jointly optimized through this objective, while the S-VAE
remains frozen.
This formulation combines temporal context modeling with
probabilistic prediction of continuous tokens.
The token processing pipeline during training and autoregressive generation during inference are detailed in the Appendix~\ref{app:b}.


%% file: Experiments/experiments.tex
We conduct extensive experiments to comprehensively evaluate HALO.
First, we describe the experimental setup (Section~\ref{sec:5.1}).
Second, we conduct in-domain and cross-domain generation experiments to evaluate the effectiveness of HALO (Section~\ref{sec:5.2}).
Third, we conduct ablation studies to investigate the contribution of each component (Section ~\ref{sec:5.3}).
Finally, we conduct efficiency analysis experiments to demonstrate the superior generation efficiency of HALO (Section ~\ref{sec:5.4}).

\subsection{Experiment Settings}\label{sec:5.1}
\paragraph{Datasets}
Our experiments are conducted on 12 datasets across four real-world domains, including:
Electricity, Solar and Wind from the \emph{energy domain};
Traffic, Taxi and Pedestrian from the \emph{transport domain};
Air Quality, Temperature and Rain from \emph{nature domain};
NN5, Fred-MD and Exchange from the \emph{economic domain}.
These real-world benchmarks are provided by GluonTS tool~\citep{gluonts}, and Monash Time Series Forecasting Repository~\citep{monash}.
For handling the datasets, we follow~\citep{timedp} and consider 3 different long sequence generation tasks, including: \{96, 168, 336\}.
More details on datasets are described in the Appendix~\ref{app:c}.


\paragraph{Baselines}
We carefully select a series of high-performance deep generative models as our baselines to achieve comprehensive evaluation.
These baselines include TimeGAN~\citep{TIMEGAN}, GT-GAN~\citep{gtgan}, TimeVAE~\citep{timevae}, TimeVQVAE~\citep{timevqvae}, TimeVQVAE-C (class-conditional version of TimeVQVAE), TimeDP~\citep{timedp}, and TimeMAR~\citep{timemar}.
Descriptions of the baselines are provided in the Appendix~\ref{app:d}.

\paragraph{Implementation Details}
All experiments are conducted on a single NVIDIA RTX 4090 GPU using Python 3.10.2 and PyTorch 2.9.1.
We train the model (both stage) for up to 100,000 iterations with a batch size of 256 and an initial learning rate of \(1\times10^{-3}\), using linear warm-up for the first 1,000 iterations followed by cosine decay.
More technical details are provided in the Appendix~\ref{app:e}.

\paragraph{Evaluation Metrics}
To evaluate generation quality, we assess both the distributional fidelity between real and synthetic time series and the preservation of their temporal characteristics. 
Following this principle, we adopt three metrics: \textbf{Maximum Mean Discrepancy} (\textbf{MMD}), \textbf{Kullback--Leibler} (\textbf{KL}) divergence, and \textbf{Marginal Distribution Difference} (\textbf{MDD}). 
\textbf{MMD} measures the global distribution discrepancy between real and synthetic samples in a kernel-induced feature space, while \textbf{KL} divergence quantifies their distributional discrepancy at the variable level. 
\textbf{MMD} further evaluates marginal distribution consistency by comparing the empirical distributions of real and synthetic data at each time step. For all three metrics, lower values indicate better generation quality.
\subsection{Results}\label{sec:5.2}

\paragraph{Evaluation of In-Domain Generation Quality}
We combine 12 datasets from 4 domains into a single multi-domain dataset for training, and then evaluate the proposed method against other advanced baselines.
To ensure a fair comparison, we follow the experimental settings of TimeDP and conduct all experiments based on its codebase.
Each run is repeated 5 times with different random seeds.
Table~\ref{tab:1} reports the mean and standard deviation of the MMD and K-L metrics for time series generation with a sequence length of 168.
Results for other sequence lengths, including the corresponding MMD scores and additional evaluation metrics, are provided in the Appendix~\ref{app:f}.

\input{Tables/table_1}

As shown in Table~\ref{tab:1}, HALO achieves leading overall
generation performance across the 12 datasets.
These improvements can be attributed in part to two
complementary design choices.
First, continuous latent modeling avoids vector quantization,
while the hyperspherical constraint stabilizes latent scales,
facilitating second-stage distribution learning.
Second, masked autoregressive modeling enables parallel token
generation conditioned on temporal context, reducing sequential
decoding steps and potentially limiting error accumulation.
Together with conditional diffusion modeling, these designs
enable HALO to generate time series with high distributional
fidelity.

\begin{figure}[h]
    \centering
    \includegraphics[width=1.0\linewidth]{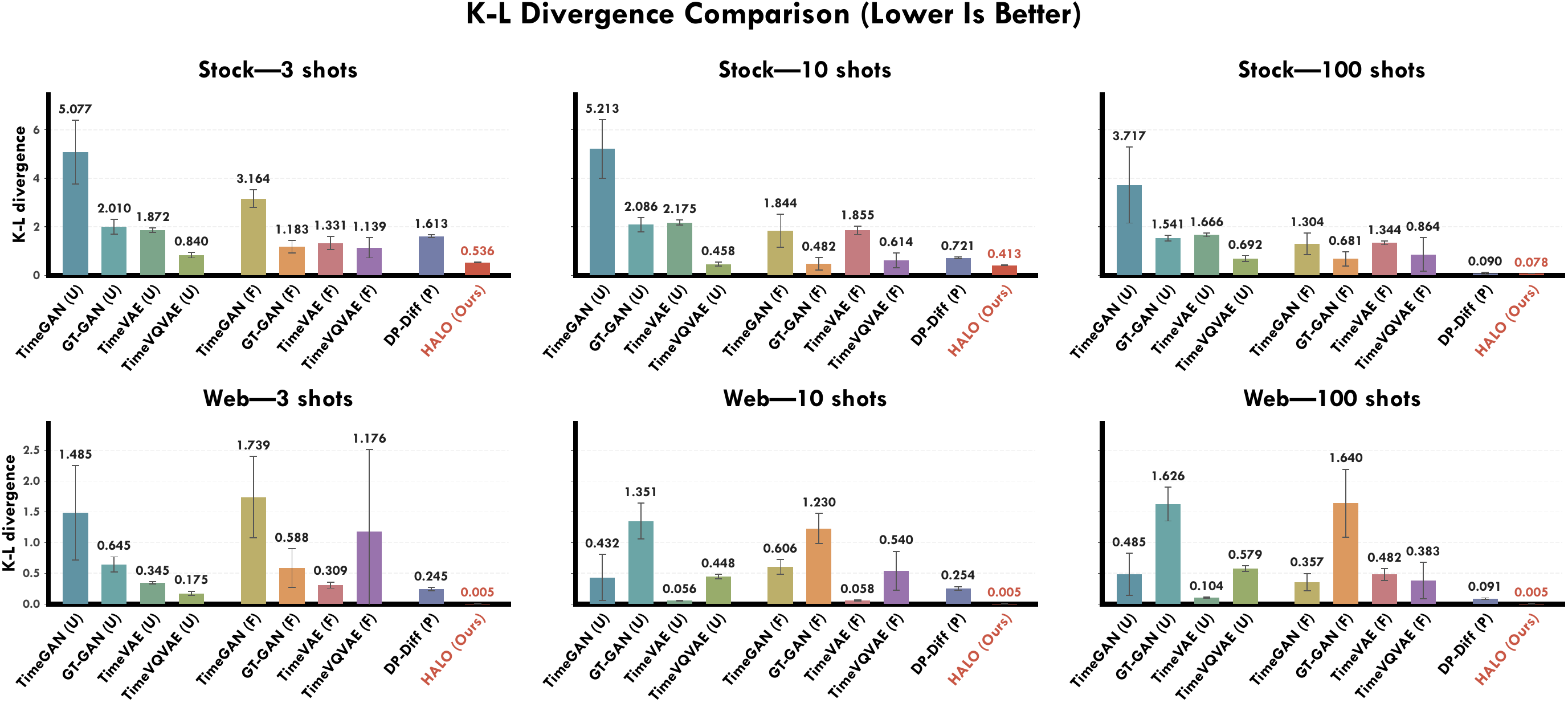}
    \vspace{-16pt}
    \caption{K-L divergence (K-L) in unseen domain settings of Stock and Web datasets for generation sequence length 168.
    U, F, and P denote unconditional generation, fine-tuning, and sample prompting, respectively.
    Results for the other metrics are provided in the Appendix~\ref{app:f}.}
    \label{fig:4}
    \vspace{-10pt}
\end{figure}

\paragraph{Evaluation of Cross-Domain Generation Quality}
Benefiting from classifier-free guidance (CFG), HALO supports unconditional generation and can be directly applied to unseen domains excluded from the multi-domain training set. 
Following~\citep{timedp}, we use 3, 10, and 100 samples as both prompts and fine-tuning data for other models, with the KL divergence results for a generation length of 168 reported in Figure~\ref{fig:4}.

As shown in Figure~\ref{fig:4}, HALO achieves state-of-the-art and robust cross-domain generation performance, substantially outperforming other unconditional baselines and even fine-tuned methods. 
Notably, HALO also maintains a clear advantage over TimeDP, which leverages samples from unseen domains as prompts, further demonstrating its strong cross-domain generalization capability and highlighting its potential for developing large-scale foundation models for time series generation.

\subsection{Ablation Studies}\label{sec:5.3}

\begin{wraptable}{r}{0.4\textwidth}
\vspace{-22pt}
\setlength{\tabcolsep}{4pt}
\renewcommand{\arraystretch}{1.0}
    \centering
    \caption{\small Maximum mean discrepancy (MMD) and K-L divergence (K-L) results for ablation study on generation sequence length 168.
    For each evaluation metric, we report the average value across all datasets.}
    \vspace{6pt}
    \label{tab:2}
    \begin{tabular}{c|c|c|c|c}
        \scalebox{0.7}{Metrics} &
        \scalebox{0.7}{HALO} & 
        \scalebox{0.7}{w/ VAE} & 
        \scalebox{0.7}{w/ AR} &
        \scalebox{0.7}{w/ MLP} \\
        \midrule[1.5pt]

        \scalebox{0.7}{MMD} &
        \scalebox{0.7}{0.078} &
        \scalebox{0.7}{0.093} &
        \scalebox{0.7}{0.099} &
        
        \scalebox{0.7}{0.274}
        \\

        \scalebox{0.7}{K-L} &
        \scalebox{0.7}{0.198} &
        \scalebox{0.7}{0.202} &
        \scalebox{0.7}{0.210} &
        \scalebox{0.7}{1.678}
        
    \vspace{-10pt}
    \end{tabular}
\end{wraptable}
To assess the contribution of each component, we independently
replace the S-VAE, masked autoregressive backbone, and conditional
diffusion head with a standard VAE (w/ VAE), a GPT-style Transformer
(w/ AR), and an MLP (w/ MLP), respectively.
As shown in Table~\ref{tab:2}, all three variants underperform the full model.
The degradation with w/ VAE suggests that reducing latent variance
and scale heterogeneity facilitates second-stage distribution modeling.
Meanwhile, w/ AR requires substantially more sequential decoding steps,
which may exacerbate error accumulation and impair generation quality.
Finally, the degradation with w/ MLP supports the effectiveness of
conditional diffusion modeling over deterministic latent regression.

\subsection{Efficiency Analysis}\label{sec:5.4}
\begin{wraptable}{r}{0.38\textwidth}
\vspace{-22pt}
\setlength{\tabcolsep}{6pt}
\renewcommand{\arraystretch}{1.0}
    \centering
    \caption{\small Generation efficiency comparison. PGM and IS denote peak GPU memory (GB) consumption and average inference time per sample (ms), respectively.}
    \vspace{6pt}
    \label{tab:3}
    \begin{tabular}{c|c|c|c}
        \scalebox{0.7}{Metrics} &
        \scalebox{0.7}{HALO} & 
        \scalebox{0.7}{TimeMAR} & 
        \scalebox{0.7}{HALO/AR} \\
        \midrule[1.5pt]

        \scalebox{0.7}{PGM} &
        \scalebox{0.7}{1.406} &
        \scalebox{0.7}{2.311} &
        \scalebox{0.7}{1.407}
        \\

        \scalebox{0.7}{IS} &
        \scalebox{0.7}{2.287} &
        \scalebox{0.7}{8.986} &
        \scalebox{0.7}{5.784}
        
    \vspace{-10pt}
    \end{tabular}
\end{wraptable}
We compare the generation efficiency of HALO, TimeMAR,
and HALO (AR) for sequences of length 168.
As shown in Table~\ref{tab:3}, HALO substantially outperforms
TimeMAR~\citep{timemar}, a state-of-the-art autoregressive baseline
for time series generation, in inference efficiency.
This advantage primarily stems from HALO's parallel decoding,
which reduces the number of sequential generation steps.
Another reason is that discrete quantization incurs additional computation for generation.


%% file: Tables/table_1.tex
\begin{table*}[!h]
\vspace{-10pt}
\centering

\setlength{\tabcolsep}{5pt}

\caption{
Maximum mean discrepancy (MMD) and K-L divergence (K-L) of in-domain generation for 168.
Best results are highlighted in bold face and second best results are underlined.
\label{tab:1}
}

\vspace{6pt}

\renewcommand{\arraystretch}{1}
\setlength{\tabcolsep}{16pt}

\resizebox{0.95\linewidth}{!}{
\begin{tabular}{cc|c|c|c|c|c|c|c|c}

\multicolumn{2}{c|}{} &
\icon \textbf{HALO} &
\textbf{TimeMAR} &
\textbf{TimeDP} &
\textbf{TimeGAN} &
\textbf{GT-GAN} &
\textbf{TimeVAE} &
\textbf{TimeVQVAE} &
\textbf{TimeVQVAE-C}
\\

\multicolumn{2}{c|}{} &
\textbf{(Ours)} &
\citeyear{timemar} &
\citeyear{timedp} &
\citeyear{TIMEGAN} &
\citeyear{gtgan} &
\citeyear{timevae} &
\citeyear{timevqvae} &
\citeyear{timevqvae}
\\

\midrule[1.5pt]

\multirow{12}{*}{
\rotatebox[origin=c]{90}{MMD}}
&
Electricity &
$\textbf{0.001}\pm0.000$
&
$\textbf{0.001}\pm0.000$
&
$\underline{0.001\pm0.001}$
&
${0.367}\pm0.255$
&
${0.254}\pm0.166$
&
${0.577}\pm0.006$
&
${0.152}\pm0.024$
&
${0.002}\pm0.001$
\\

&
Solar &
$\underline{0.033\pm0.000}$
&
$\textbf{0.032}\pm0.000$
&
${0.041}\pm0.011$
&
${0.628}\pm0.053$
&
${0.578}\pm0.039$
&
${0.353}\pm0.014$
&
${0.437}\pm0.020$
&
${0.058}\pm0.005$
\\

&
Wind &
$\textbf{0.007}\pm0.000$
&
${0.023}\pm0.001$
&
${0.025}\pm0.017$
&
${0.213}\pm0.017$
&
${0.170}\pm0.040$
&
${0.170}\pm0.004$
&
${0.131}\pm0.014$
&
$\underline{0.018\pm0.007}$
\\

&
Traffic &
$\underline{0.068\pm0.001}$
&
$\textbf{0.068}\pm0.000$
&
${0.083\pm0.034}$
&
${0.567}\pm0.057$
&
${0.538}\pm0.078$
&
${0.218}\pm0.007$
&
${0.213}\pm0.016$
&
${0.089}\pm0.001$
\\

&
Taxi &
$\textbf{0.083}\pm0.001$
&
${0.150}\pm0.002$
&
$\underline{0.095\pm0.023}$
&
${0.275}\pm0.054$
&
${0.319}\pm0.032$
&
${0.139}\pm0.007$
&
${0.128}\pm0.004$
&
${0.109}\pm0.014$
\\

&
Pedestrian &
${0.071}\pm0.001$
&
${0.067}\pm0.001$
&
$\textbf{0.044}\pm0.020$
&
${0.090}\pm0.030$
&
${0.112}\pm0.019$
&
${0.065}\pm0.002$
&
${0.067}\pm0.007$
&
$\underline{0.058\pm0.002}$
\\

&
Air &
$\textbf{0.010}\pm0.000$
&
${0.031}\pm0.001$
&
$\underline{0.011\pm0.003}$
&
${0.120}\pm0.045$
&
${0.211}\pm0.041$
&
${0.089}\pm0.016$
&
${0.028}\pm0.002$
&
${0.041}\pm0.008$
\\

&
Temperature &
${0.247}\pm0.003$
&
$\underline{0.235\pm0.005}$
&
$\textbf{0.219}\pm0.022$
&
${0.926}\pm0.042$
&
${0.809}\pm0.081$
&
${1.002}\pm0.014$
&
${0.323}\pm0.008$
&
${0.259}\pm0.043$
\\

&
Rain &
$\textbf{0.025}\pm0.000$
&
${0.074}\pm0.002$
&
$\underline{0.057\pm0.039}$
&
${0.329}\pm0.285$
&
${0.111}\pm0.109$
&
${0.292}\pm0.019$
&
${0.074}\pm0.007$
&
${0.080}\pm0.004$
\\

&
NN5 &
${0.199}\pm0.002$
&
$\textbf{0.161}\pm0.002$
&
$\underline{0.164\pm0.010}$
&
${0.874}\pm0.088$
&
${0.632}\pm0.074$
&
${0.821}\pm0.061$
&
${0.327}\pm0.012$
&
${0.243}\pm0.041$
\\

&
Fred-MD &
$\textbf{0.001}\pm0.000$
&
$\underline{0.002}\pm0.000$
&
${0.002\pm0.001}$
&
${0.043}\pm0.021$
&
${0.133}\pm0.102$
&
${0.059}\pm0.008$
&
${0.008}\pm0.002$
&
${0.005}\pm0.002$
\\

&
Exchange &
${0.184}\pm0.003$
&
$\textbf{0.125}\pm0.003$
&
$\underline{0.151\pm0.024}$
&
${0.530}\pm0.154$
&
${0.475}\pm0.116$
&
${0.543}\pm0.149$
&
${0.342}\pm0.050$
&
${0.233}\pm0.107$
\\

\midrule[1.5pt]

\multirow{12}{*}{
\rotatebox[origin=c]{90}{K-L}}
&
Electricity &
$\textbf{0.002}\pm0.000$
&
$\underline{0.012\pm0.003}$
&
$\underline{0.012\pm0.016}$
&
${0.488}\pm0.175$
&
${0.407}\pm0.079$
&
${0.734}\pm0.023$
&
${0.280}\pm0.051$
&
${0.027}\pm0.015$
\\

&
Solar &
${0.019}\pm0.000$
&
$\textbf{0.012}\pm0.000$
&
$\underline{0.016\pm0.005}$
&
${0.612}\pm0.447$
&
${0.120}\pm0.041$
&
${0.260}\pm0.016$
&
${0.865}\pm0.108$
&
${0.234}\pm0.062$
\\

&
Wind &
$\textbf{0.063}\pm0.002$
&
$\underline{0.107\pm0.006}$
&
${0.152}\pm0.034$
&
${1.924}\pm1.233$
&
$\underline{0.107\pm0.016}$
&
${0.484}\pm0.015$
&
${0.483}\pm0.066$
&
${0.183}\pm0.047$
\\

&
Traffic &
$\textbf{0.004}\pm0.000$
&
${0.014}\pm0.001$
&
$\underline{0.009\pm0.003}$
&
${1.305}\pm0.320$
&
${1.409}\pm0.251$
&
${0.211}\pm0.014$
&
${0.178}\pm0.026$
&
${0.016}\pm0.003$
\\

&
Taxi &
$\textbf{0.008}\pm0.000$
&
${0.173}\pm0.012$
&
$\underline{0.011\pm0.004}$
&
${0.650}\pm0.180$
&
${0.950}\pm0.197$
&
${0.110}\pm0.020$
&
${0.110}\pm0.026$
&
${0.038}\pm0.010$
\\

&
Pedestrian &
$\underline{0.035\pm0.000}$
&
$\underline{0.035\pm0.001}$
&
$\textbf{0.014}\pm0.010$
&
${0.417}\pm0.181$
&
${0.411}\pm0.096$
&
${0.065}\pm0.005$
&
${0.405}\pm0.051$
&
${0.039}\pm0.008$
\\

&
Air &
$\textbf{0.020}\pm0.001$
&
${0.070}\pm0.005$
&
$\underline{0.027\pm0.016}$
&
${0.348}\pm0.093$
&
${0.578}\pm0.049$
&
${0.164}\pm0.012$
&
${0.054}\pm0.012$
&
${0.093}\pm0.025$
\\

&
Temperature &
$\textbf{0.158}\pm0.006$
&
${0.220}\pm0.011$
&
$\underline{0.171\pm0.073}$
&
${8.892}\pm2.681$
&
${3.174}\pm2.685$
&
${2.183}\pm0.110$
&
${0.735}\pm0.066$
&
${0.379}\pm0.110$
\\

&
Rain &
$\textbf{0.006}\pm0.000$
&
${0.081}\pm0.002$
&
$\underline{0.013\pm0.012}$
&
${0.506}\pm0.174$
&
${0.432}\pm0.099$
&
${0.160}\pm0.022$
&
${0.047}\pm0.018$
&
${0.065}\pm0.018$
\\

&
NN5 &
$\underline{0.092\pm0.002}$
&
${0.094}\pm0.004$
&
$\textbf{0.054}\pm0.014$
&
${4.928}\pm4.112$
&
${1.386}\pm0.520$
&
${1.337}\pm0.220$
&
${1.063}\pm0.274$
&
${0.220}\pm0.151$
\\

&
Fred-MD &
${0.228}\pm0.010$
&
$\underline{0.207\pm0.008}$
&
$\textbf{0.203}\pm0.035$
&
${0.512}\pm0.290$
&
${0.380}\pm0.070$
&
${0.346}\pm0.041$
&
${0.831}\pm0.077$
&
${1.118}\pm0.276$
\\

&
Exchange &
$\textbf{1.742}\pm0.043$
&
${1.887}\pm0.020$
&
$\underline{1.866\pm0.132}$
&
${8.861}\pm3.397$
&
${7.201}\pm4.380$
&
${10.404}\pm1.434$
&
${5.052}\pm1.385$
&
${8.475}\pm3.056$
\\

\midrule[1.5pt]

\multicolumn{2}{c|}{$1^{st}$ Counts}
&
\textbf{14}
&
\underline{6}
&
5
&
0
&
0
&
0
&
0
&
0
\\

\end{tabular}
}

\end{table*}

%% file: Conclusions/conclusions.tex
In this paper, we propose HALO, a two-stage generative framework that combines continuous latent
representations with efficient masked autoregressive modeling.
Its hyperspherical VAE avoids discrete vector quantization
and alleviates latent variance and scale heterogeneity,
while its masked autoregressive model integrates temporal
context and conditional diffusion to enable parallel generation
with fewer sequential inference steps.
Experiments across 12 datasets demonstrate strong generation
quality and improved inference efficiency, with additional
evaluations showing promising generalization to unseen domains.
These results support continuous latent modeling as an
effective approach to high-quality and efficient time series
generation.

%% file: Appendix/appendix.tex
\section{Token Processing Path in Masked Context Modeling}\label{app:b}
\begin{wrapfigure}{r}{0.52\textwidth}
\vspace{-14pt}
    \begin{center}
    \includegraphics[width=1.0\linewidth]{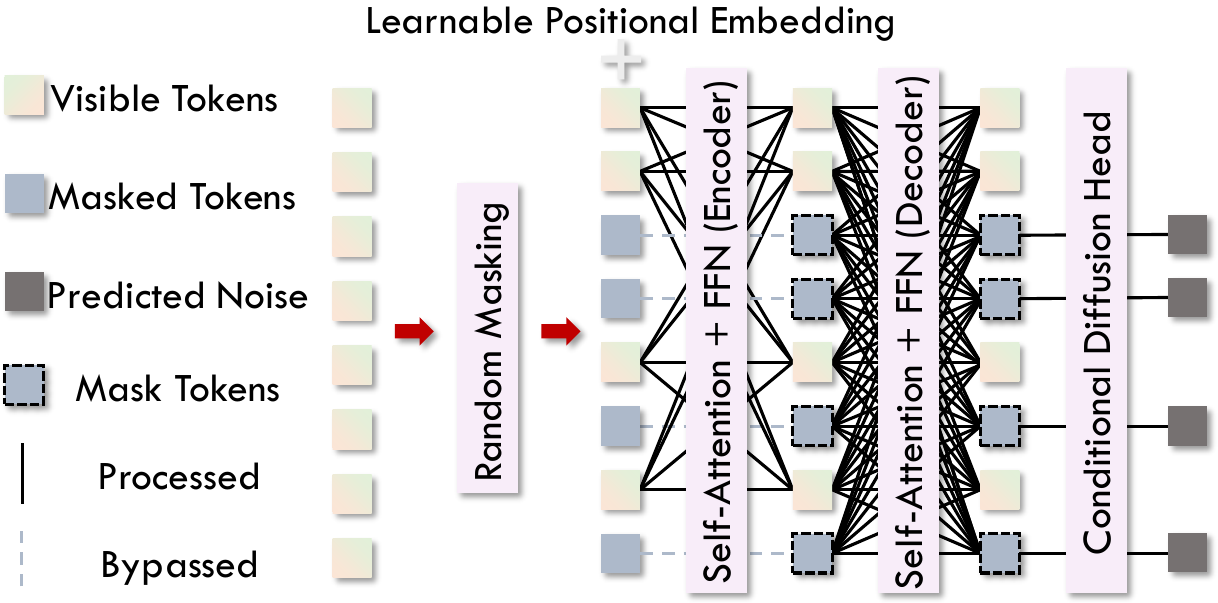} \\
    \end{center}
    \caption{\small
    Illustration of the complete token processing flow in masked autoregressive modeling.
    \label{fig:5}}
    \vspace{-10pt}
\end{wrapfigure}
\paragraph{Training}
Figure~\ref{fig:5} illustrates the complete token processing pipeline in masked autoregressive modeling, during training phase.
The encoder processes only the unmasked visible tokens, while the decoder jointly processes the encoded visible tokens and the learnable mask tokens.
Consequently, each mask token is conditioned on information propagated from the visible tokens, including their temporal positional information.
By conditioning the denoising of ground-truth masked tokens on mask tokens enriched with visible-token information, including temporal positional information, we effectively realize masked autoregressive modeling that \emph{predicts unknown tokens conditioned on known tokens}.

\input{algorithm/algorithm}
\paragraph{Inference}
Algorithm~\ref{alg:halo_inference} summarizes the inference
procedure of HALO.
Starting from a fully masked latent sequence, HALO progressively
generates groups of tokens according to a randomly sampled order
and a cosine unmasking schedule.
At each iteration, the context network conditions on all
previously generated tokens, and the conditional diffusion head
samples the selected tokens in parallel.
Here, $\tau_t$ denotes the diffusion timestep associated with
sampling step $t$, and \textsc{ReverseStep} performs one DDPM
reverse update under the selected sampling schedule, without
injecting additional noise at the final step.
Each completed token is projected onto the fixed-radius
hypersphere before being incorporated into the visible context.
Finally, the S-VAE decoder maps the completed latent sequence
back to the time series.
For unconditional generation, the domain label is replaced
with the learned null condition.
For the comparative experiments, the number of inference steps is set to \(M/6\).

\section{Dataset Description}\label{app:c}
We provide detailed descriptions into the datasets used in model training this paper:
\begin{itemize}
    \item 
    \textbf{Electricity}. This dataset contains hourly electricity consumption measurements, in kilowatts (kW), from 321 clients over the period from 2012 to 2014. The data were originally collected from the UCI Machine Learning Repository.

    \item 
    \textbf{Solar}. This dataset contains 137 time series recording hourly solar power production in the state of Alabama, USA, in 2006.

    \item 
    \textbf{Wind}. This dataset consists of a single long time series of wind power production, measured in megawatts (MW) at 4-second intervals starting from August 1, 2019. The data were obtained from the online platform of the Australian Energy Market Operator (AEMO).

    \item 
    \textbf{Traffic}. This dataset contains 15 months of daily traffic data, comprising 440 daily records. It describes the occupancy rates, ranging from 0 to 1, of different freeway lanes in the San Francisco Bay Area over time.

    \item 
    \textbf{Taxi}. This dataset contains spatio-temporal traffic time series of New York City taxi rides collected from 1,214 locations at 30-minute intervals during January 2015 and January 2016.
    
    \item 
    \textbf{Pedestrian}. This dataset contains hourly pedestrian counts collected by 66 sensors across Melbourne, Australia, starting from May 2009. The original dataset is continuously updated as new observations become available. The version used in our experiments contains records up to April 30, 2020.
    
    \item 
    \textbf{Air Quality}. This dataset was originally used in the KDD Cup 2018 forecasting competition. It contains hourly air-quality measurements collected from 59 monitoring stations in Beijing (35 stations) and London (24 stations) between January 1, 2017 and March 31, 2018. The measurements include PM2.5, PM10, NO\(_2\), CO, O\(_3\), and SO\(_2\), resulting in 270 hourly time series categorized by city, station, and measurement type. The original data contain missing values. Leading missing values are filled with zeros, while the remaining missing values are imputed using the last observation carried forward (LOCF) strategy.
    
    \item 
    \textbf{Temperature}. This dataset contains 32,072 daily records of temperature observations and rainfall forecasts collected by the Australian Bureau of Meteorology from 422 weather stations across Australia between May 2, 2015 and April 26, 2017. Missing values in the original dataset are replaced with zeros. In our experiments, we use the mean temperature variable.
    
    \item 
    \textbf{Rain}. This dataset is derived from the same source as the Temperature dataset. We extract the rainfall variable to construct the corresponding time series.
    
    \item 
    \textbf{NN5}. This dataset was originally used in the NN5 forecasting competition and contains 111 time series from the banking domain. Each series records daily cash withdrawals from an automated teller machine (ATM) in the United Kingdom. Missing values are imputed using the median value computed over the corresponding day of the week across the entire series.
    
    \item 
    \textbf{FRED-MD}. This dataset contains 107 monthly time series of macroeconomic indicators provided by the Federal Reserve Bank and obtained from the FRED-MD database. Following common practice in the literature, the series are appropriately differenced and log-transformed.
    
    \item 
    \textbf{Exchange}. This dataset contains daily exchange rates among eight currencies.

\end{itemize}

Moreover, we also describe the dataset used in unseen data generation experiment:
\begin{itemize}
    \item 

    \textbf{Stock}. This dataset contains the daily stock prices of Google (GOOG), which is listed on the NASDAQ stock exchange.
    
    \item 

    \textbf{Web}. This dataset was originally used in the Kaggle Wikipedia Web Traffic Forecasting competition. It contains 145,063 daily time series recording the number of page views for a collection of Wikipedia pages from July 1, 2015 to September 10, 2017. Missing values in the original dataset are replaced with zeros.
\end{itemize}

\section{Baseline Details}\label{app:d}

We compare HALO with the following time series generation baselines,
covering adversarial, variational, diffusion-based, and discrete-token
generative approaches.
Table~\ref{tab:4} summarizes the code repositories of the baseline methods.

\begin{table*}[h]
    \centering
    \setlength{\tabcolsep}{2pt}
    \vspace{-16pt}
    \caption{\small Code repositories of the baseline methods.
    TimeVQVAE-C shares the implementation of TimeVQVAE.}
    \vspace{6pt}
    \label{tab:4}
    \small
    \begin{tabular}{lc}
        \toprule[1.5pt]
        \textbf{Baseline} & \textbf{Code Repository} \\
        \midrule
        TimeMAR
        & \url{https://github.com/decisionintelligence/TimeMAR} \\
        \midrule
        TimeDP
        & \url{https://github.com/microsoft/TimeCraft/tree/main/TimeDP} \\
        \midrule
        TimeGAN
        & \url{https://github.com/jsyoon0823/TimeGAN} \\
        \midrule
        GT-GAN
        & \url{https://github.com/Jinsung-Jeon/GT-GAN} \\
        \midrule
        TimeVAE
        & \url{https://github.com/abudesai/timeVAE} \\
        \midrule
        TimeVQVAE
        & \url{https://github.com/ML4ITS/TimeVQVAE} \\
        \midrule
        TimeVQVAE-C
        & \url{https://github.com/ML4ITS/TimeVQVAE} \\
        
        \bottomrule[1.5pt]
    \end{tabular}
\end{table*}

\paragraph{TimeMAR~\citep{timemar}.}
TimeMAR is a two-stage generative framework that combines multi-scale
discrete representations with autoregressive modeling.
It employs a dual-path VQ-VAE to encode decomposed trend and seasonal
components into discrete tokens at multiple temporal resolutions,
followed by a Transformer that generates these tokens in a
coarse-to-fine manner.
Coarse seasonal representations further guide the reconstruction
of fine-grained seasonal patterns.
Following the experimental protocol of~\citep{timedp}, we adapt it to use domain labels as prompts.

\paragraph{TimeDP~\citep{timedp}.}
TimeDP is a diffusion-based framework for multi-domain time series
generation.
It learns a shared set of semantic prototypes to represent temporal
patterns and employs a prototype assignment module to construct
domain prompts from input time series.
These prompts condition the diffusion model, enabling domain-specific
generation using a small set of examples from the target domain.

\paragraph{TimeGAN~\citep{TIMEGAN}.}
TimeGAN combines adversarial learning with a supervised temporal
objective in a learned embedding space.
Its embedding and recovery networks provide latent representations
for generation, while the supervised objective encourages the model
to preserve temporal dependencies between consecutive latent states.

\paragraph{GT-GAN~\citep{gtgan}.}
GT-GAN is an adversarial framework designed to synthesize both
regularly and irregularly sampled time series.
It integrates neural ordinary and controlled differential equations
with continuous-time flow processes to capture temporal dynamics
under different sampling patterns.

\paragraph{TimeVAE~\citep{timevae}.}
TimeVAE is a variational autoencoder for multivariate time series
generation.
It combines continuous latent representations with an interpretable
decoder that explicitly models level, trend, and seasonal components,
allowing generated sequences to reflect these temporal structures.

\paragraph{TimeVQVAE~\citep{timevqvae}.}
TimeVQVAE adopts a two-stage framework based on vector quantization
and bidirectional prior modeling.
It separates time-frequency representations into low- and
high-frequency components and encodes them into discrete tokens.
Bidirectional Transformer priors learn the token distributions
through masked modeling, enabling generation via iterative
masked decoding.

\paragraph{TimeVQVAE-C~\citep{timevqvae,timedp}.}
TimeVQVAE-C denotes the class-conditional variant of TimeVQVAE.
It retains the time-frequency tokenization framework while
conditioning the bidirectional token prior on class labels,
enabling the generation of time series associated with a
specified class.


\section{More Implementation Details}\label{app:e}

We provide additional implementation details for the S-VAE (Stage-1) and masked autoregressive model (Stage-2).

\paragraph{S-VAE}

Our S-VAE adopts a hybrid convolutional--Transformer
encoder--decoder architecture.
The encoder contains two residual downsampling blocks
followed by three Transformer blocks, while the decoder
contains three Transformer blocks followed by two residual
upsampling blocks.
Each Transformer block uses a hidden dimension of 512
and eight attention heads.
The encoder predicts the parameters of a Power Spherical
posterior over 16-dimensional ($d$) latent tokens, which are
constrained to a hypersphere of radius 4 ($\sqrt{d}$).
The temporal downsampling factor is set to 4 ($s$).

\paragraph{Masked Autoregressive Model}
Our masked autoregressive model consists of a random masking
strategy, a Transformer-based context modeling network,
and a conditional diffusion head.
The context encoder and decoder each contain two Transformer
blocks with a hidden dimension of 256 ($D$) and 8 attention heads.
Learnable positional embeddings encode temporal positions,
while masked positions are represented by learnable mask tokens
in the decoder.
The conditional diffusion head employs two residual MLP blocks
with a hidden dimension of 512, conditioned on contextual
representations and diffusion timesteps through adaptive
layer normalization.
The maximum number of diffusion sampling steps is set to 100 ($t$).
The minimum masking ratio is set to 0.5 ($r$).

\section{Complete Experimental Results}\label{app:f}

\begin{figure}[h]
    \centering
    \includegraphics[width=1.0\linewidth]{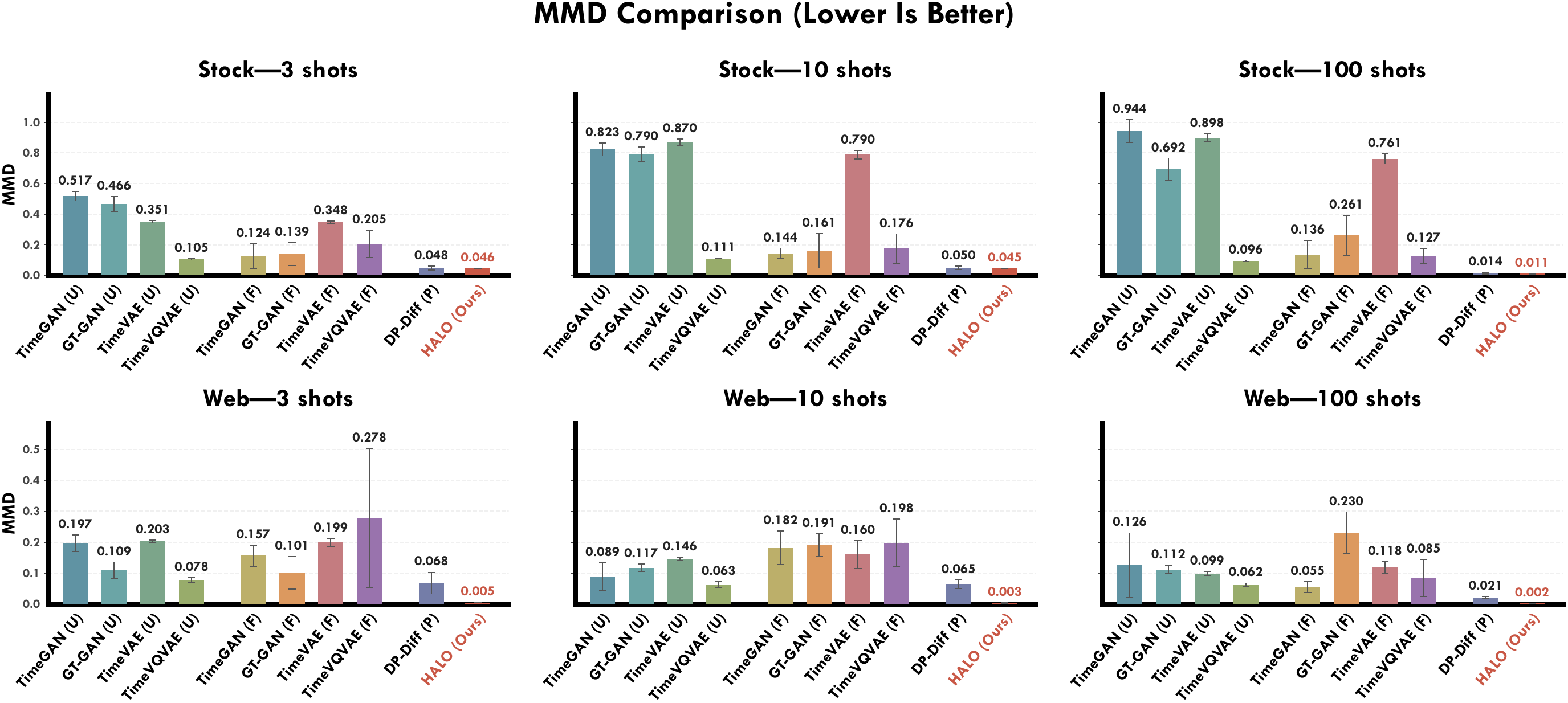} \\
    \vspace{6pt}
    \includegraphics[width=1.0\linewidth]{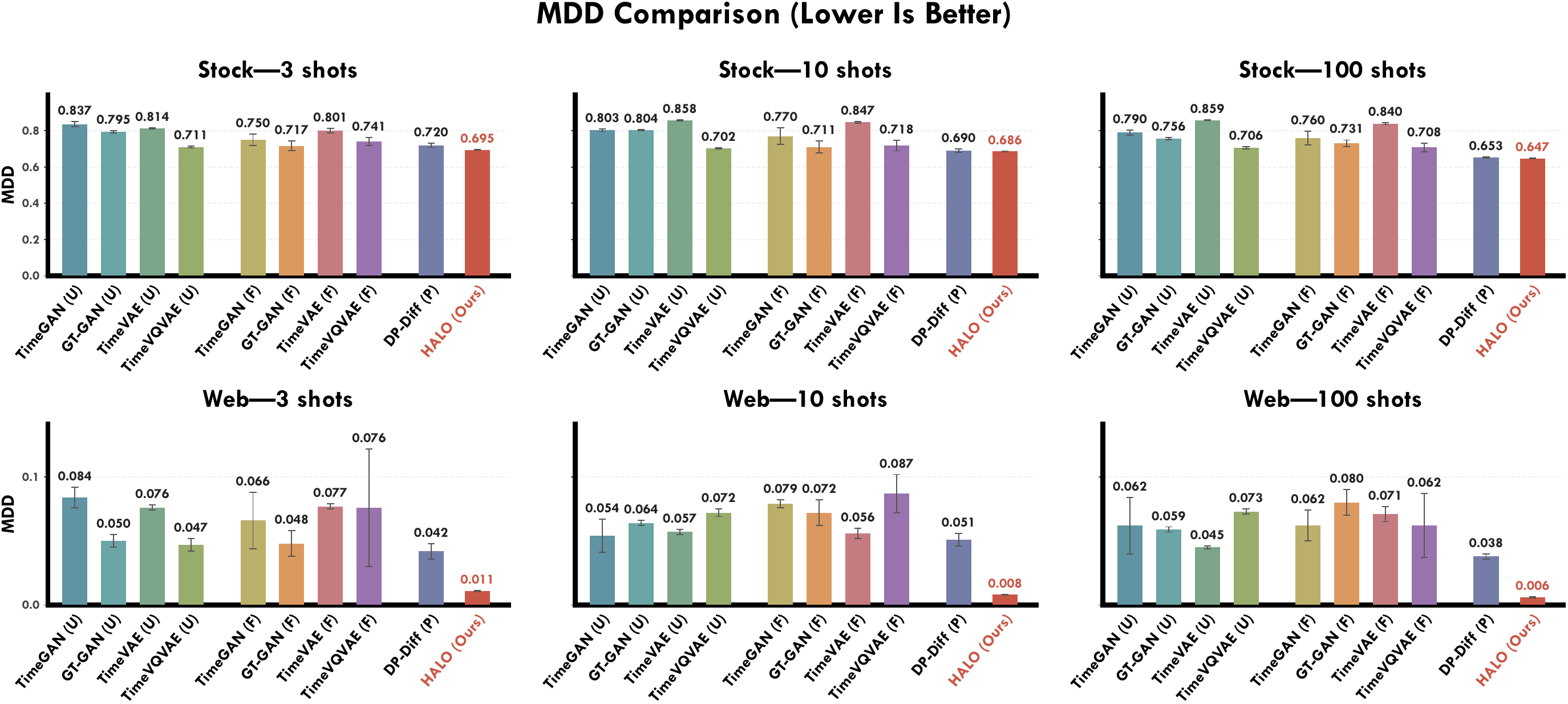}
    
    \caption{Maximum mean discrepancy (MMD) and Marginal distribution
    distance (MDD) in unseen domain settings of Stock and Web datasets for generation sequence length 168.
    U, F, and P denote unconditional generation, fine-tuning, and sample prompting, respectively.}
    \label{fig:6}
\end{figure}

Table~\ref{tab:5} reports the MDD results corresponding to Table~\ref{tab:1} in the main text.
In addition, the MMD and MDD results for the cross-domain generation experiments are presented in Figures~\ref{fig:6}, respectively.
Results for the remaining generation sequence lengths are provided in Tables~\ref{tab:6} to ~\ref{tab:7}.

\begin{table*}[h]
\centering
\setlength{\tabcolsep}{5pt}
\caption{Marginal distribution distance (MDD) of in-domain generation for 168. Best results are highlighted in bold face and second best results are underlined.\label{tab:5}}
\vspace{6pt}
\renewcommand{\arraystretch}{1}
\setlength{\tabcolsep}{16pt}
\resizebox{0.95\linewidth}{!}{
\begin{tabular}{cc|c|c|c|c|c|c|c|c}
\multicolumn{2}{c|}{} &
\icon \textbf{HALO} &
\textbf{TimeMAR} &
\textbf{TimeDP} &
\textbf{TimeGAN} &
\textbf{GT-GAN} &
\textbf{TimeVAE} &
\textbf{TimeVQVAE} &
\textbf{TimeVQVAE-C}
\\
\multicolumn{2}{c|}{} &
\textbf{(Ours)} &
\citeyear{timemar} &
\citeyear{timedp} &
\citeyear{TIMEGAN} &
\citeyear{gtgan} &
\citeyear{timevae} &
\citeyear{timevqvae} &
\citeyear{timevqvae}
\\
\midrule[1.5pt]

\multirow{12}{*}{\rotatebox[origin=c]{90}{MDD}}
& Electricity
& $\textbf{0.002}\pm0.000$
& $\underline{0.003\pm0.000}$
& $0.005\pm0.002$
& $0.075\pm0.035$
& $0.047\pm0.008$
& $0.098\pm0.003$
& $0.067\pm0.004$
& $0.005\pm0.001$
\\
& Solar
& $63.027\pm0.192$
& $\underline{39.173\pm0.226}$
& $56.414\pm21.890$
& $70.334\pm11.928$
& $83.855\pm3.100$
& $\textbf{16.721}\pm0.041$
& $57.401\pm0.041$
& $59.043\pm0.440$
\\
& Wind
& $\textbf{0.058}\pm0.002$
& $\underline{0.072\pm0.002}$
& $0.084\pm0.009$
& $0.226\pm0.061$
& $0.138\pm0.015$
& $0.201\pm0.004$
& $0.159\pm0.011$
& $0.085\pm0.008$
\\
& Traffic
& $0.051\pm0.000$
& $\textbf{0.048}\pm0.001$
& $\underline{0.049\pm0.004}$
& $0.149\pm0.018$
& $0.153\pm0.001$
& $0.110\pm0.001$
& $0.119\pm0.005$
& $0.053\pm0.002$
\\
& Taxi
& $\textbf{0.080}\pm0.001$
& $0.095\pm0.000$
& $\underline{0.081\pm0.008}$
& $0.104\pm0.010$
& $0.109\pm0.004$
& $0.094\pm0.001$
& $0.096\pm0.004$
& $0.084\pm0.004$
\\
& Pedestrian
& $0.089\pm0.000$
& $0.088\pm0.001$
& $\textbf{0.071}\pm0.012$
& $0.096\pm0.024$
& $0.097\pm0.006$
& $0.086\pm0.002$
& $0.143\pm0.007$
& $\underline{0.079\pm0.002}$
\\
& Air
& $\textbf{0.038}\pm0.001$
& $0.046\pm0.001$
& $\underline{0.042\pm0.002}$
& $0.139\pm0.025$
& $0.171\pm0.011$
& $0.085\pm0.001$
& $0.092\pm0.008$
& $0.058\pm0.006$
\\
& Temperature
& $\textbf{0.142}\pm0.001$
& $\underline{0.143\pm0.002}$
& $\textbf{0.142}\pm0.010$
& $0.189\pm0.009$
& $0.208\pm0.012$
& $0.259\pm0.004$
& $0.191\pm0.004$
& $0.156\pm0.012$
\\
& Rain
& $0.079\pm0.000$
& $\textbf{0.061}\pm0.000$
& $\underline{0.067\pm0.015}$
& $0.228\pm0.090$
& $0.142\pm0.020$
& $0.228\pm0.008$
& $0.177\pm0.016$
& $0.104\pm0.006$
\\
& NN5
& $0.150\pm0.001$
& $\textbf{0.134}\pm0.000$
& $\underline{0.140\pm0.005}$
& $0.295\pm0.021$
& $0.240\pm0.014$
& $0.304\pm0.010$
& $0.220\pm0.012$
& $0.175\pm0.011$
\\
& Fred-MD
& $0.022\pm0.000$
& $\textbf{0.018}\pm0.000$
& $\underline{0.021\pm0.002}$
& $0.079\pm0.016$
& $0.098\pm0.024$
& $0.104\pm0.009$
& $0.126\pm0.009$
& $0.076\pm0.025$
\\
& Exchange
& $0.358\pm0.001$
& $\textbf{0.326}\pm0.003$
& $0.358\pm0.010$
& $0.351\pm0.098$
& $\underline{0.345\pm0.018}$
& $0.442\pm0.020$
& $0.515\pm0.021$
& $0.486\pm0.040$
\\

\end{tabular}}
\end{table*}

    


\input{Tables/full_96}
\clearpage
\input{Tables/full_336}
\section{Additional Experiments}\label{app:g}
\paragraph{Visualization of Synthesized Time Series}
\begin{figure}[h]
    \centering
    \includegraphics[width=0.95\linewidth]{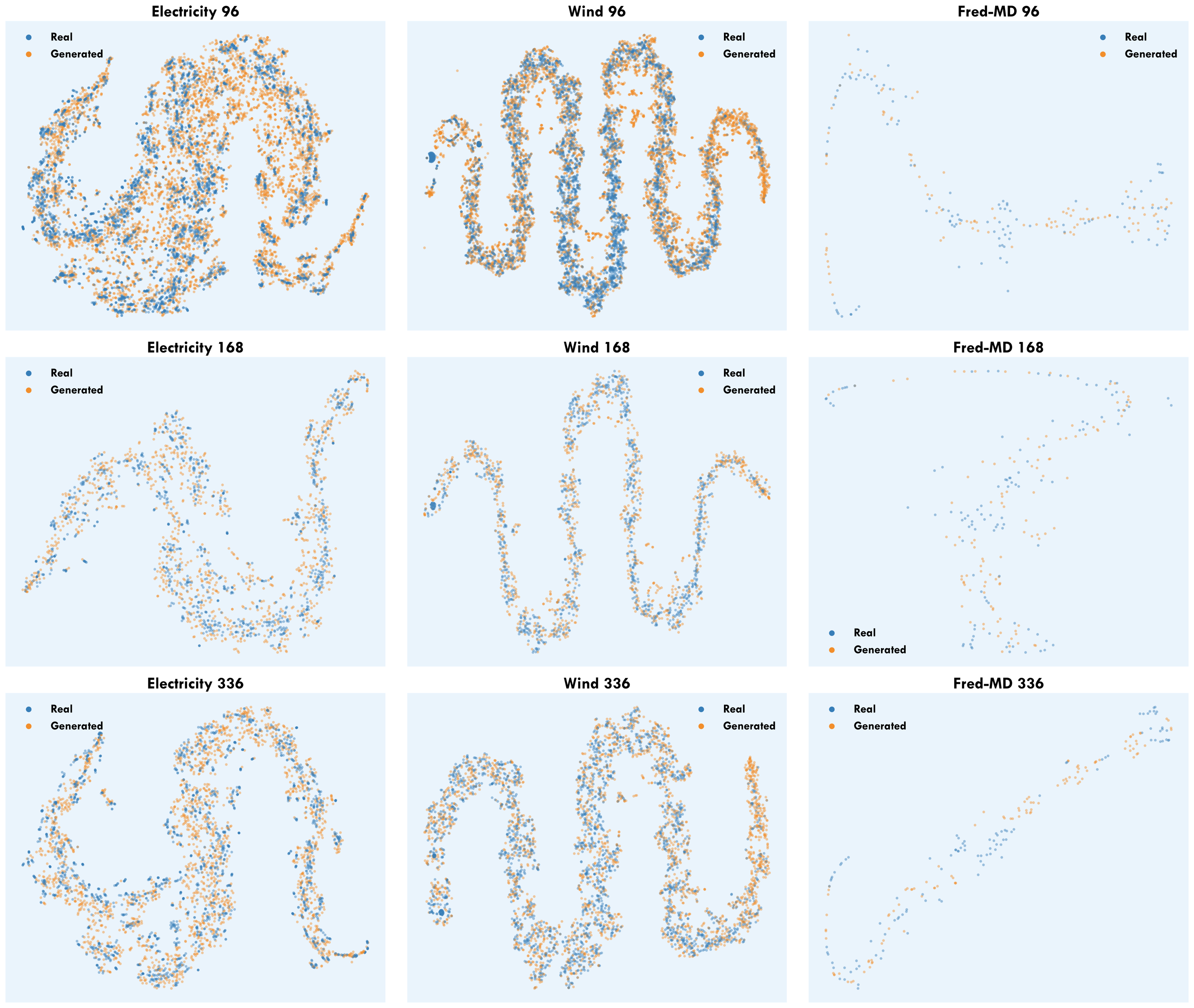}
    \caption{\small Visualization of synthesized time series by t-SNE across different sequence lengths.}
    \label{fig:7}
\end{figure}
We employ t-SNE to visualize the distributional similarity between real and generated samples, providing a qualitative assessment of generation fidelity, as shown in Figure~\ref{fig:7}.
We observe a strong distributional alignment between the synthesized time-series samples and the real data.
This demonstrates that HALO is capable of generating synthetic time series that faithfully preserve both the statistical properties and temporal structures of the original data.

\paragraph{Sensitivity Analysis of Autoregressive Generation Rounds}

To examine how the number of autoregressive generation rounds ($K$ in Algorithm~\ref{alg:halo_inference}) affects generation quality, we conduct a sensitivity analysis.
Table~\ref{tab:8} reports the results with varying numbers of inference rounds. 
The results suggest that a more fine-grained autoregressive generation process may lead to higher-quality synthetic time series, potentially because it better preserves temporal dependencies during generation.
This observation further supports our motivation: effectively balancing generation efficiency and temporal correlation modeling is key to achieving a desirable trade-off among generation stability, accuracy, and computational efficiency.

\begin{table*}[!h]
\centering
\setlength{\tabcolsep}{16pt}
\caption{Generation quality of HALO with different numbers of inference iterations $K$ for sequence length 168. Results are reported as mean $\pm$ sample standard deviation over five random seeds. Best results are highlighted in bold face and second best results are underlined.\label{tab:8}}
\vspace{6pt}
\renewcommand{\arraystretch}{1}
\resizebox{0.95\linewidth}{!}{
\begin{tabular}{cc|c|c|c|c}
\multicolumn{2}{c|}{} &
$K=1$ &
$K=2$ &
$K=4$ &
$K=12$
\\
\midrule[1.5pt]

\multirow{3}{*}{\rotatebox[origin=c]{90}{MMD}}
& Electricity
& $0.003\pm0.000$
& $0.003\pm0.000$
& $\textbf{0.001}\pm0.000$
& $\textbf{0.001}\pm0.000$
\\
& Solar
& $0.041\pm0.000$
& $\textbf{0.033}\pm0.001$
& $\underline{0.034\pm0.000}$
& $\textbf{0.033}\pm0.001$
\\
& Traffic
& $0.087\pm0.001$
& $0.083\pm0.001$
& $\underline{0.072\pm0.001}$
& $\textbf{0.065}\pm0.001$
\\
\midrule[1.5pt]

\multirow{3}{*}{\rotatebox[origin=c]{90}{K-L}}
& Electricity
& $0.025\pm0.001$
& $0.059\pm0.002$
& $\underline{0.020\pm0.002}$
& $\textbf{0.002}\pm0.001$
\\
& Solar
& $0.143\pm0.007$
& $0.092\pm0.008$
& $\underline{0.035\pm0.002}$
& $\textbf{0.013}\pm0.000$
\\
& Traffic
& $0.013\pm0.000$
& $\underline{0.006\pm0.000}$
& $\underline{0.006\pm0.000}$
& $\textbf{0.002}\pm0.000$
\\
\midrule[1.5pt]

\multirow{3}{*}{\rotatebox[origin=c]{90}{MDD}}
& Electricity
& $0.008\pm0.000$
& $0.009\pm0.000$
& $\underline{0.006\pm0.000}$
& $\textbf{0.003}\pm0.000$
\\
& Solar
& $82.188\pm0.157$
& $75.515\pm0.152$
& $\underline{68.501\pm0.215}$
& $\textbf{58.702}\pm0.369$
\\
& Traffic
& $0.056\pm0.000$
& $0.056\pm0.000$
& $\underline{0.055\pm0.000}$
& $\textbf{0.047}\pm0.000$
\\
\end{tabular}}
\end{table*}

\section{Additional Related Work}

The time series analysis community has extensively explored
forecasting~\citep{TFB}, anomaly detection~\citep{tab},
classification~\citep{classsurvey}, and
imputation~\citep{missingsurvey}.
Research on these tasks has advanced temporal representation
learning, dependency modeling, and evaluation methodologies,
providing broader context for time series generation.
A central challenge in time series modeling is capturing
heterogeneous temporal patterns and complex channel
interactions~\citep{duet,graphpatchformer,timemosaic,jifine,DAG}.
Beyond these, refining learning objectives
offers a promising direction to enhance representation learning
in time series models~\citep{cvloss,dbloss}.
This perspective also extends to time series generation,
where carefully designed objectives can encourage the
preservation of temporal structure and improve generation
fidelity~\citep{diffusionts}.
Complementing these advances, HALO focuses on the geometry
of continuous latent representations and the generation
process, combining hyperspherical constraints with masked
autoregressive modeling to improve generation quality
and efficiency.

%% file: algorithm/algorithm.tex

\begin{algorithm}[h]
\caption{Inference Procedure of HALO}
\label{alg:halo_inference}
\small
\begin{algorithmic}[1]
\Require Trained context network $f_\theta$, noise predictor
$\epsilon_\omega$, and S-VAE decoder $\operatorname{Decoder}$
\Require Number of latent tokens $M$, latent dimension $d$,
and radius $R=\sqrt{d}$
\Require Autoregressive iterations $1\leq K\leq M$,
diffusion sampling steps $S=100$, and sampling schedule
$\{\tau_t\}_{t=1}^{S}$
\Require Optional domain label $C$
($C=\varnothing$ for the learned null condition)
\Ensure Generated time series $\widehat{X}$

\State Initialize $Z=[z_1,\ldots,z_M]
\gets\mathbf{0}_{M\times d}$
\State Initialize $\mathcal{M}\gets\{1,\ldots,M\}$
and $\mathcal{V}\gets\varnothing$
\State Sample a random permutation
$\pi=(\pi_1,\ldots,\pi_M)$ of $\{1,\ldots,M\}$

\For{$k=1,\ldots,K$}
    \If{$k=K$}
        \State $m_k\gets 0$
        \Comment{Generate all remaining tokens}
    \Else
        \State $m_k\gets
        \left\lfloor
        M\cos\left(\frac{\pi k}{2K}\right)
        \right\rfloor$
        \State $m_k\gets
        \max\{1,\min\{|\mathcal{M}|-1,m_k\}\}$
    \EndIf

    \State $\mathcal{M}_{\mathrm{next}}
    \gets\{\pi_j:1\leq j\leq m_k\}$
    \State $\mathcal{P}_k
    \gets\mathcal{M}\setminus\mathcal{M}_{\mathrm{next}}$
    \Comment{Positions generated in this iteration}

    \State $H\gets f_\theta(Z_{\mathcal{V}},\mathcal{M},C)$
    \Comment{Context from previously generated tokens}

    \ForAll{$i\in\mathcal{P}_k$ \textbf{in parallel}}
        \State Sample
        $z_i^{(S)}\sim\mathcal{N}(0,I_d)$
        \For{$t=S,S-1,\ldots,1$}
            \State $\widehat{\epsilon}_i^{(t)}
            \gets\epsilon_\omega(z_i^{(t)},\tau_t,H_i)$
            \State $z_i^{(t-1)}
            \gets\Call{ReverseStep}{
            z_i^{(t)},\widehat{\epsilon}_i^{(t)},t}$
        \EndFor
        \State $z_i\gets
        R\,z_i^{(0)}/\|z_i^{(0)}\|_2$
        \Comment{Fixed-radius projection}
    \EndFor

    \State $\mathcal{V}\gets\mathcal{V}\cup\mathcal{P}_k$
    \State $\mathcal{M}\gets\mathcal{M}_{\mathrm{next}}$
\EndFor

\State $\widehat{X}\gets\operatorname{Decoder}(Z)$
\State \Return $\widehat{X}$
\end{algorithmic}
\end{algorithm}

%% file: Tables/full_96.tex
\begin{table*}[!h]
\centering
\setlength{\tabcolsep}{5pt}
\caption{Maximum mean discrepancy (MMD), K-L divergence (K-L), and marginal distribution distance (MDD) of in-domain generation for 96. Best results are highlighted in bold face and second best results are underlined.\label{tab:6}}
\vspace{6pt}
\renewcommand{\arraystretch}{1}
\setlength{\tabcolsep}{16pt}
\resizebox{0.95\linewidth}{!}{
\begin{tabular}{cc|c|c|c|c|c|c|c|c}
\multicolumn{2}{c|}{} &
\icon \textbf{HALO} &
\textbf{TimeMAR} &
\textbf{TimeDP} &
\textbf{TimeGAN} &
\textbf{GT-GAN} &
\textbf{TimeVAE} &
\textbf{TimeVQVAE} &
\textbf{TimeVQVAE-C}
\\
\multicolumn{2}{c|}{} &
\textbf{(Ours)} &
\citeyear{timemar} &
\citeyear{timedp} &
\citeyear{TIMEGAN} &
\citeyear{gtgan} &
\citeyear{timevae} &
\citeyear{timevqvae} &
\citeyear{timevqvae}
\\
\midrule[1.5pt]

\multirow{12}{*}{\rotatebox[origin=c]{90}{MMD}}
& Electricity
& $\textbf{0.001}\pm0.000$
& $\textbf{0.001}\pm0.000$
& $\underline{0.002\pm0.001}$
& $0.401\pm0.219$
& $0.317\pm0.311$
& $0.648\pm0.012$
& $0.153\pm0.023$
& $\underline{0.002\pm0.000}$
\\
& Solar
& $\underline{0.017\pm0.000}$
& $\textbf{0.012}\pm0.000$
& $0.019\pm0.001$
& $0.584\pm0.056$
& $0.634\pm0.119$
& $0.369\pm0.012$
& $0.439\pm0.021$
& $0.035\pm0.008$
\\
& Wind
& $0.041\pm0.005$
& $0.031\pm0.002$
& $\underline{0.029\pm0.015}$
& $0.195\pm0.030$
& $0.217\pm0.134$
& $0.179\pm0.002$
& $0.129\pm0.016$
& $\textbf{0.013}\pm0.007$
\\
& Traffic
& $\underline{0.019\pm0.000}$
& $0.023\pm0.000$
& $\textbf{0.018}\pm0.005$
& $0.456\pm0.100$
& $0.513\pm0.073$
& $0.162\pm0.005$
& $0.189\pm0.014$
& $0.027\pm0.002$
\\
& Taxi
& $\textbf{0.053}\pm0.000$
& $0.060\pm0.001$
& $\underline{0.055\pm0.007}$
& $0.197\pm0.057$
& $0.328\pm0.071$
& $0.060\pm0.002$
& $0.089\pm0.005$
& $0.061\pm0.002$
\\
& Pedestrian
& $0.064\pm0.001$
& $0.056\pm0.001$
& $\textbf{0.040}\pm0.010$
& $0.068\pm0.025$
& $0.159\pm0.149$
& $0.064\pm0.001$
& $0.066\pm0.011$
& $\underline{0.049\pm0.005}$
\\
& Air
& $\textbf{0.008}\pm0.000$
& $\textbf{0.008}\pm0.000$
& $\underline{0.010\pm0.006}$
& $0.103\pm0.046$
& $0.244\pm0.160$
& $0.085\pm0.011$
& $0.022\pm0.003$
& $0.036\pm0.007$
\\
& Temperature
& $\textbf{0.157}\pm0.003$
& $\underline{0.177\pm0.005}$
& $0.178\pm0.047$
& $0.827\pm0.092$
& $0.859\pm0.109$
& $0.958\pm0.012$
& $0.413\pm0.034$
& $0.205\pm0.033$
\\
& Rain
& $\textbf{0.023}\pm0.001$
& $0.056\pm0.001$
& $\underline{0.044\pm0.027}$
& $0.285\pm0.199$
& $0.166\pm0.203$
& $0.327\pm0.020$
& $0.068\pm0.008$
& $0.072\pm0.001$
\\
& NN5
& $\underline{0.213\pm0.001}$
& $\textbf{0.212}\pm0.002$
& $0.234\pm0.047$
& $0.888\pm0.091$
& $0.706\pm0.116$
& $0.875\pm0.053$
& $0.321\pm0.008$
& $0.241\pm0.011$
\\
& Fred-MD
& $\underline{0.002\pm0.000}$
& $\textbf{0.001}\pm0.000$
& $0.005\pm0.002$
& $0.055\pm0.022$
& $0.150\pm0.184$
& $0.046\pm0.013$
& $0.004\pm0.001$
& $0.006\pm0.001$
\\
& Exchange
& $\textbf{0.117}\pm0.004$
& $\underline{0.141\pm0.003}$
& $0.343\pm0.210$
& $0.449\pm0.018$
& $0.567\pm0.114$
& $0.592\pm0.124$
& $0.337\pm0.056$
& $0.224\pm0.028$
\\
\midrule[1.5pt]

\multirow{12}{*}{\rotatebox[origin=c]{90}{K-L}}
& Electricity
& $\textbf{0.002}\pm0.001$
& $0.026\pm0.004$
& $\underline{0.015\pm0.017}$
& $0.552\pm0.241$
& $0.447\pm0.147$
& $0.790\pm0.018$
& $0.245\pm0.039$
& $0.031\pm0.005$
\\
& Solar
& $0.021\pm0.002$
& $\underline{0.014\pm0.000}$
& $\textbf{0.010}\pm0.002$
& $0.922\pm0.953$
& $0.117\pm0.047$
& $0.288\pm0.015$
& $0.812\pm0.077$
& $0.151\pm0.114$
\\
& Wind
& $0.166\pm0.009$
& $\textbf{0.122}\pm0.005$
& $0.182\pm0.029$
& $2.288\pm1.135$
& $\underline{0.164\pm0.071}$
& $0.527\pm0.009$
& $0.510\pm0.112$
& $0.284\pm0.115$
\\
& Traffic
& $\textbf{0.004}\pm0.000$
& $0.019\pm0.000$
& $\underline{0.010\pm0.006}$
& $1.241\pm0.757$
& $1.198\pm0.340$
& $0.205\pm0.008$
& $0.192\pm0.016$
& $0.014\pm0.004$
\\
& Taxi
& $\textbf{0.004}\pm0.001$
& $0.045\pm0.002$
& $\underline{0.010\pm0.007}$
& $0.550\pm0.333$
& $0.801\pm0.270$
& $0.121\pm0.003$
& $0.102\pm0.017$
& $0.027\pm0.007$
\\
& Pedestrian
& $\underline{0.025\pm0.000}$
& $0.040\pm0.003$
& $\textbf{0.012}\pm0.005$
& $0.458\pm0.368$
& $0.327\pm0.148$
& $0.074\pm0.004$
& $0.373\pm0.027$
& $0.034\pm0.006$
\\
& Air
& $\textbf{0.021}\pm0.002$
& $\underline{0.022\pm0.001}$
& $0.025\pm0.011$
& $0.317\pm0.102$
& $0.510\pm0.147$
& $0.168\pm0.012$
& $0.061\pm0.012$
& $0.096\pm0.019$
\\
& Temperature
& $\textbf{0.301}\pm0.007$
& $\underline{0.388\pm0.013}$
& $0.392\pm0.097$
& $11.298\pm1.212$
& $5.830\pm4.888$
& $1.737\pm0.068$
& $1.135\pm0.110$
& $0.567\pm0.050$
\\
& Rain
& $\underline{0.028\pm0.001}$
& $0.096\pm0.002$
& $\textbf{0.022}\pm0.009$
& $0.549\pm0.257$
& $0.481\pm0.146$
& $0.237\pm0.025$
& $0.055\pm0.021$
& $0.091\pm0.010$
\\
& NN5
& $\textbf{0.067}\pm0.003$
& $0.071\pm0.002$
& $\underline{0.069\pm0.033}$
& $5.829\pm6.120$
& $2.709\pm2.080$
& $1.537\pm0.224$
& $1.038\pm0.168$
& $0.103\pm0.072$
\\
& Fred-MD
& $\textbf{0.288}\pm0.002$
& $\underline{0.328\pm0.008}$
& $1.183\pm0.280$
& $0.628\pm0.335$
& $0.486\pm0.108$
& $0.441\pm0.063$
& $0.813\pm0.093$
& $1.240\pm0.232$
\\
& Exchange
& $\textbf{1.911}\pm0.115$
& $\underline{2.660\pm0.116}$
& $14.336\pm2.652$
& $6.445\pm2.210$
& $10.170\pm6.381$
& $9.937\pm4.322$
& $3.740\pm1.261$
& $4.408\pm0.700$
\\
\midrule[1.5pt]

\multirow{12}{*}{\rotatebox[origin=c]{90}{MDD}}
& Electricity
& $\textbf{0.002}\pm0.000$
& $\underline{0.003\pm0.000}$
& $0.005\pm0.002$
& $0.070\pm0.020$
& $0.046\pm0.011$
& $0.082\pm0.002$
& $0.058\pm0.002$
& $0.005\pm0.001$
\\
& Solar
& $64.900\pm0.081$
& $72.076\pm0.169$
& $77.545\pm18.384$
& $60.819\pm9.001$
& $77.198\pm5.522$
& $\textbf{47.123}\pm0.114$
& $\underline{50.619\pm0.099}$
& $53.220\pm0.832$
\\
& Wind
& $0.085\pm0.003$
& $\underline{0.076\pm0.003}$
& $0.089\pm0.012$
& $0.207\pm0.054$
& $0.135\pm0.035$
& $0.208\pm0.002$
& $0.159\pm0.009$
& $\textbf{0.069}\pm0.014$
\\
& Traffic
& $\underline{0.023\pm0.000}$
& $\textbf{0.021}\pm0.000$
& $0.024\pm0.003$
& $0.109\pm0.018$
& $0.122\pm0.009$
& $0.083\pm0.001$
& $0.089\pm0.002$
& $0.028\pm0.002$
\\
& Taxi
& $\underline{0.043\pm0.000}$
& $0.054\pm0.000$
& $\underline{0.043\pm0.004}$
& $0.072\pm0.008$
& $0.092\pm0.010$
& $0.064\pm0.001$
& $0.070\pm0.003$
& $\textbf{0.040}\pm0.002$
\\
& Pedestrian
& $0.069\pm0.000$
& $\underline{0.065\pm0.001}$
& $\textbf{0.058}\pm0.007$
& $0.074\pm0.018$
& $0.088\pm0.025$
& $0.073\pm0.001$
& $0.116\pm0.003$
& $\textbf{0.058}\pm0.003$
\\
& Air
& $\underline{0.028\pm0.001}$
& $\textbf{0.024}\pm0.001$
& $0.031\pm0.007$
& $0.091\pm0.021$
& $0.145\pm0.022$
& $0.070\pm0.002$
& $0.066\pm0.001$
& $0.056\pm0.007$
\\
& Temperature
& $\textbf{0.117}\pm0.001$
& $\underline{0.127\pm0.002}$
& $0.135\pm0.016$
& $0.154\pm0.004$
& $0.185\pm0.007$
& $0.196\pm0.001$
& $0.198\pm0.006$
& $0.143\pm0.010$
\\
& Rain
& $0.040\pm0.000$
& $\textbf{0.033}\pm0.000$
& $\underline{0.039\pm0.006}$
& $0.192\pm0.059$
& $0.113\pm0.011$
& $0.196\pm0.003$
& $0.134\pm0.009$
& $0.047\pm0.012$
\\
& NN5
& $\textbf{0.148}\pm0.001$
& $\underline{0.150\pm0.000}$
& $0.175\pm0.016$
& $0.270\pm0.014$
& $0.225\pm0.017$
& $0.268\pm0.006$
& $0.196\pm0.007$
& $0.158\pm0.004$
\\
& Fred-MD
& $\textbf{0.014}\pm0.000$
& $\textbf{0.014}\pm0.000$
& $\underline{0.023\pm0.005}$
& $0.063\pm0.009$
& $0.076\pm0.023$
& $0.055\pm0.012$
& $0.083\pm0.004$
& $0.045\pm0.017$
\\
& Exchange
& $\textbf{0.312}\pm0.003$
& $0.340\pm0.002$
& $0.402\pm0.083$
& $0.342\pm0.041$
& $\underline{0.334\pm0.024}$
& $0.425\pm0.027$
& $0.503\pm0.032$
& $0.426\pm0.017$
\\

\end{tabular}}
\end{table*}

%% file: Tables/full_336.tex
\begin{table*}[!h]
\centering
\setlength{\tabcolsep}{5pt}
\caption{Maximum mean discrepancy (MMD), K-L divergence (K-L), and marginal distribution distance (MDD) of in-domain generation for 336. Best results are highlighted in bold face and second best results are underlined. TimeMAR could not be evaluated due to out-of-memory (OOM).\label{tab:7}}
\vspace{6pt}
\renewcommand{\arraystretch}{1}
\setlength{\tabcolsep}{16pt}
\resizebox{0.95\linewidth}{!}{
\begin{tabular}{cc|c|c|c|c|c|c|c|c}
\multicolumn{2}{c|}{} &
\icon \textbf{HALO} &
\textbf{TimeMAR} &
\textbf{TimeDP} &
\textbf{TimeGAN} &
\textbf{GT-GAN} &
\textbf{TimeVAE} &
\textbf{TimeVQVAE} &
\textbf{TimeVQVAE-C}
\\
\multicolumn{2}{c|}{} &
\textbf{(Ours)} &
\citeyear{timemar} &
\citeyear{timedp} &
\citeyear{TIMEGAN} &
\citeyear{gtgan} &
\citeyear{timevae} &
\citeyear{timevqvae} &
\citeyear{timevqvae}
\\
\midrule[1.5pt]

\multirow{12}{*}{\rotatebox[origin=c]{90}{MMD}}
& Electricity
& $\textbf{0.000}\pm0.000$
& OOM
& $\underline{0.001\pm0.002}$
& $0.320\pm0.197$
& $0.347\pm0.524$
& $0.504\pm0.012$
& $0.135\pm0.018$
& $0.003\pm0.000$
\\
& Solar
& $\textbf{0.066}\pm0.001$
& OOM
& $\underline{0.074\pm0.012}$
& $0.646\pm0.032$
& $0.735\pm0.045$
& $0.400\pm0.015$
& $0.493\pm0.024$
& $0.081\pm0.006$
\\
& Wind
& $0.027\pm0.001$
& OOM
& $\underline{0.024\pm0.003}$
& $0.208\pm0.095$
& $0.221\pm0.010$
& $0.145\pm0.005$
& $0.139\pm0.015$
& $\textbf{0.014}\pm0.007$
\\
& Traffic
& $\underline{0.081\pm0.001}$
& OOM
& $\textbf{0.077}\pm0.009$
& $0.518\pm0.028$
& $0.704\pm0.101$
& $0.235\pm0.011$
& $0.233\pm0.017$
& $0.091\pm0.005$
\\
& Taxi
& $0.214\pm0.002$
& OOM
& $\underline{0.189\pm0.034}$
& $0.329\pm0.011$
& $0.390\pm0.078$
& $\textbf{0.166}\pm0.002$
& $0.202\pm0.004$
& $\underline{0.189\pm0.007}$
\\
& Pedestrian
& $\underline{0.055\pm0.001}$
& OOM
& $\textbf{0.035}\pm0.008$
& $0.140\pm0.067$
& $0.158\pm0.019$
& $0.061\pm0.003$
& $0.074\pm0.007$
& $0.067\pm0.008$
\\
& Air
& $\textbf{0.016}\pm0.000$
& OOM
& $\underline{0.018\pm0.005}$
& $0.213\pm0.037$
& $0.261\pm0.062$
& $0.075\pm0.005$
& $0.038\pm0.003$
& $0.053\pm0.007$
\\
& Temperature
& $0.199\pm0.002$
& OOM
& $\textbf{0.185}\pm0.021$
& $0.914\pm0.031$
& $0.934\pm0.051$
& $0.976\pm0.022$
& $0.318\pm0.008$
& $\underline{0.197\pm0.014}$
\\
& Rain
& $\textbf{0.035}\pm0.000$
& OOM
& $0.096\pm0.086$
& $0.151\pm0.108$
& $0.198\pm0.295$
& $0.135\pm0.018$
& $\underline{0.066\pm0.002}$
& $0.081\pm0.005$
\\
& NN5
& $\textbf{0.187}\pm0.001$
& OOM
& $\underline{0.191\pm0.016}$
& $0.663\pm0.027$
& $0.951\pm0.069$
& $0.806\pm0.086$
& $0.333\pm0.009$
& $0.260\pm0.049$
\\
& Fred-MD
& $\textbf{0.003}\pm0.000$
& OOM
& $\underline{0.006\pm0.004}$
& $0.156\pm0.097$
& $0.063\pm0.006$
& $0.052\pm0.020$
& $0.010\pm0.006$
& $0.007\pm0.001$
\\
& Exchange
& $\textbf{0.126}\pm0.003$
& OOM
& $0.683\pm0.217$
& $0.724\pm0.130$
& $0.553\pm0.238$
& $0.598\pm0.166$
& $0.318\pm0.077$
& $\underline{0.275\pm0.198}$
\\
\midrule[1.5pt]

\multirow{12}{*}{\rotatebox[origin=c]{90}{K-L}}
& Electricity
& $\textbf{0.005}\pm0.001$
& OOM
& $\underline{0.016\pm0.016}$
& $0.396\pm0.108$
& $1.169\pm2.029$
& $0.719\pm0.020$
& $0.257\pm0.023$
& $0.042\pm0.033$
\\
& Solar
& $\textbf{0.016}\pm0.001$
& OOM
& $\underline{0.017\pm0.006}$
& $0.179\pm0.062$
& $2.150\pm2.179$
& $0.249\pm0.012$
& $0.870\pm0.110$
& $0.201\pm0.017$
\\
& Wind
& $\textbf{0.098}\pm0.005$
& OOM
& $0.158\pm0.032$
& $0.169\pm0.064$
& $4.264\pm2.790$
& $0.373\pm0.019$
& $0.495\pm0.043$
& $\underline{0.121\pm0.037}$
\\
& Traffic
& $\textbf{0.004}\pm0.000$
& OOM
& $\underline{0.008\pm0.004}$
& $1.443\pm0.354$
& $2.992\pm0.592$
& $0.241\pm0.012$
& $0.210\pm0.034$
& $0.023\pm0.006$
\\
& Taxi
& $0.107\pm0.001$
& OOM
& $\textbf{0.076}\pm0.034$
& $0.937\pm0.151$
& $1.461\pm0.367$
& $0.160\pm0.015$
& $0.230\pm0.026$
& $\underline{0.094\pm0.007}$
\\
& Pedestrian
& $\underline{0.014\pm0.000}$
& OOM
& $\textbf{0.009}\pm0.003$
& $0.452\pm0.131$
& $1.544\pm0.908$
& $0.044\pm0.007$
& $0.407\pm0.069$
& $0.053\pm0.017$
\\
& Air
& $\textbf{0.009}\pm0.001$
& OOM
& $\underline{0.017\pm0.010}$
& $1.031\pm0.145$
& $1.481\pm0.553$
& $0.180\pm0.015$
& $0.121\pm0.028$
& $0.079\pm0.016$
\\
& Temperature
& $\textbf{0.024}\pm0.001$
& OOM
& $\underline{0.037\pm0.016}$
& $2.479\pm1.867$
& $9.139\pm2.247$
& $1.726\pm0.122$
& $1.119\pm0.179$
& $0.224\pm0.123$
\\
& Rain
& $\textbf{0.008}\pm0.000$
& OOM
& $\underline{0.011\pm0.002}$
& $0.388\pm0.112$
& $0.271\pm0.046$
& $0.028\pm0.005$
& $0.025\pm0.013$
& $0.028\pm0.005$
\\
& NN5
& $\textbf{0.046}\pm0.002$
& OOM
& $\underline{0.075\pm0.034}$
& $1.418\pm0.122$
& $8.449\pm5.473$
& $1.233\pm0.256$
& $0.836\pm0.250$
& $0.180\pm0.181$
\\
& Fred-MD
& $\textbf{0.203}\pm0.012$
& OOM
& $0.815\pm0.630$
& $0.309\pm0.145$
& $0.568\pm0.321$
& $\underline{0.212\pm0.047}$
& $0.586\pm0.161$
& $1.099\pm0.020$
\\
& Exchange
& $\textbf{1.971}\pm0.063$
& OOM
& $18.426\pm4.493$
& $14.116\pm3.798$
& $13.875\pm2.965$
& $10.670\pm5.335$
& $7.755\pm2.566$
& $\underline{7.651\pm3.841}$
\\
\midrule[1.5pt]

\multirow{12}{*}{\rotatebox[origin=c]{90}{MDD}}
& Electricity
& $\textbf{0.003}\pm0.000$
& OOM
& $\underline{0.006\pm0.004}$
& $0.048\pm0.009$
& $0.061\pm0.058$
& $0.099\pm0.002$
& $0.069\pm0.003$
& $0.008\pm0.003$
\\
& Solar
& $\underline{40.105\pm0.113}$
& OOM
& $49.170\pm21.121$
& $82.860\pm4.163$
& $74.403\pm18.119$
& $\textbf{13.755}\pm0.077$
& $54.065\pm0.183$
& $55.878\pm0.249$
\\
& Wind
& $\textbf{0.068}\pm0.003$
& OOM
& $0.091\pm0.007$
& $0.146\pm0.028$
& $0.251\pm0.060$
& $0.172\pm0.004$
& $0.168\pm0.009$
& $\underline{0.070\pm0.009}$
\\
& Traffic
& $\textbf{0.059}\pm0.000$
& OOM
& $\underline{0.066\pm0.002}$
& $0.191\pm0.005$
& $0.206\pm0.031$
& $0.134\pm0.001$
& $0.158\pm0.005$
& $0.072\pm0.005$
\\
& Taxi
& $\underline{0.684\pm0.001}$
& OOM
& $\textbf{0.680}\pm0.021$
& $1.073\pm0.162$
& $0.986\pm0.126$
& $0.764\pm0.003$
& $0.750\pm0.003$
& $0.707\pm0.001$
\\
& Pedestrian
& $0.090\pm0.001$
& OOM
& $\textbf{0.074}\pm0.006$
& $0.121\pm0.027$
& $0.153\pm0.070$
& $\underline{0.086\pm0.002}$
& $0.161\pm0.007$
& $0.098\pm0.006$
\\
& Air
& $\textbf{0.042}\pm0.000$
& OOM
& $\underline{0.049\pm0.004}$
& $0.157\pm0.002$
& $0.166\pm0.039$
& $0.078\pm0.002$
& $0.099\pm0.005$
& $0.056\pm0.005$
\\
& Temperature
& $\textbf{0.140}\pm0.001$
& OOM
& $\underline{0.153\pm0.010}$
& $0.230\pm0.007$
& $0.218\pm0.013$
& $0.341\pm0.005$
& $0.204\pm0.009$
& $0.155\pm0.006$
\\
& Rain
& $\underline{0.108\pm0.000}$
& OOM
& $\textbf{0.100}\pm0.007$
& $0.144\pm0.011$
& $0.238\pm0.184$
& $0.210\pm0.014$
& $0.222\pm0.009$
& $0.138\pm0.014$
\\
& NN5
& $\textbf{1.033}\pm0.003$
& OOM
& $1.079\pm0.016$
& $1.128\pm0.017$
& $1.199\pm0.024$
& $1.209\pm0.021$
& $1.109\pm0.006$
& $\underline{1.074\pm0.012}$
\\
& Fred-MD
& $\textbf{0.025}\pm0.000$
& OOM
& $\underline{0.031\pm0.007}$
& $0.106\pm0.038$
& $0.085\pm0.006$
& $0.093\pm0.035$
& $0.128\pm0.016$
& $0.060\pm0.025$
\\
& Exchange
& $0.386\pm0.001$
& OOM
& $0.510\pm0.131$
& $\textbf{0.337}\pm0.023$
& $\underline{0.383\pm0.091}$
& $0.443\pm0.027$
& $0.537\pm0.036$
& $0.452\pm0.048$
\\

\end{tabular}}
\end{table*}